\documentclass[journal]{IEEEtran}
\usepackage{multirow,booktabs,color,soul}
\usepackage{amsmath,rotating}
\usepackage{subfig} 
\usepackage{epsfig}
\usepackage{algorithmic}
\usepackage{algorithm}
\usepackage{multirow,booktabs,color,soul,threeparttable,rotating}
\graphicspath{{Figure/}}
\usepackage{url}
\definecolor{hl}{rgb}{0.75,0.75,0.75}
\sethlcolor{hl}
\usepackage{hyperref}

\begin{document}
%
\title{RelativeNAS: Relative Neural Architecture\\ Search via Slow-Fast Learning}
%
%
%
\author{Hao Tan, Ran Cheng, \textit{Senior Member, IEEE}, Shihua Huang, Cheng He, \textit{Member, IEEE},\\ 
Changxiao Qiu, Fan Yang, Ping Luo%
\thanks{H. Tan, R. Cheng, S. Huang, and C. He are with University Key Laboratory of Evolving Intelligent Systems of Guangdong Province, Department of Computer Science and Engineering, Southern University of Science and Technology, Shenzhen 518055, P.R. China. \textcolor{black}{E-mails: tanbox@live.com, ranchengcn@gmail.com, shihuahuang95@gmail.com, chenghehust@gmail.com.} (\textit{Corresponding author: Ran Cheng}) }%
\thanks{C. Qiu and F. Yang are with Hisilicon Research Department, Huawei Technologies Co.,Ltd.
 Shenzhen 518055, P.R. China. \textcolor{black}{E-mails: \{qiuchangxiao, yangfan74\}@huawei.com.}}%
 \thanks{P. Luo is with Department of Computer Science, The University of Hong Kong, Hong Kong, P.R. China.  \textcolor{black}{E-mail: pluo.lhi@gmail.com}.}
 }

\markboth{IEEE Transactions on Neural Networks and Learning Systems,~Vol.~xx, No.~xx, xxxx}%
{Hao Tan \MakeLowercase{\textit{et al.}}: RelativeNAS: Relative Neural Architecture Search via Slow-Fast Learning}

\maketitle

\begin{abstract}
Despite the remarkable successes of Convolutional Neural Networks (CNNs) in computer vision, it is time-consuming and error-prone to manually design a CNN.
Among various Neural Architecture Search (NAS) methods that are motivated to automate designs of high-performance CNNs, the differentiable NAS and population-based NAS are attracting increasing interests due to their unique characters.
To benefit from the merits while overcoming the deficiencies of both, this work proposes a novel NAS method, RelativeNAS.
As the key to efficient search, RelativeNAS performs joint learning between fast-learners (\emph{i.e.} decoded networks with relatively lower loss value) and slow-learners in a pairwise manner.
Moreover, since RelativeNAS only requires low-fidelity performance estimation to distinguish each pair of fast-learner and slow-learner, it saves certain computation costs for training the candidate architectures.
The proposed RelativeNAS brings several unique advantages: 
(1) it achieves state-of-the-art performances on ImageNet with top-$\mathbf{1}$ error rate of $\mathbf{24.88\%}$, \emph{i.e.}  outperforming DARTS and AmoebaNet-B by $\mathbf{1.82\%}$ and $\mathbf{1.12\%}$ respectively;
(2) it spends only nine hours with a single 1080Ti GPU to obtain the discovered cells, \emph{i.e.}  $\mathbf{3.75\times}$ and $\mathbf{7875\times}$ faster than DARTS and AmoebaNet respectively; 
(3) it provides that the discovered cells obtained on CIFAR-$\mathbf{10}$ can be directly transferred to object detection, semantic segmentation, and keypoint detection, yielding competitive results of $\mathbf{73.1\%}$ mAP on PASCAL VOC, $\mathbf{78.7\%}$ mIoU on Cityscapes, and $\mathbf{68.5\%}$ AP on MSCOCO, respectively.
The implementation of RelativeNAS is available at \href{https://github.com/EMI-Group/RelativeNAS}{https://github.com/EMI-Group/RelativeNAS}.
\end{abstract}


\IEEEpeerreviewmaketitle

\section{Introduction}\label{sec:introduction}

\begin{figure*}[!htb]
     \centering
     \includegraphics[width=1\linewidth]{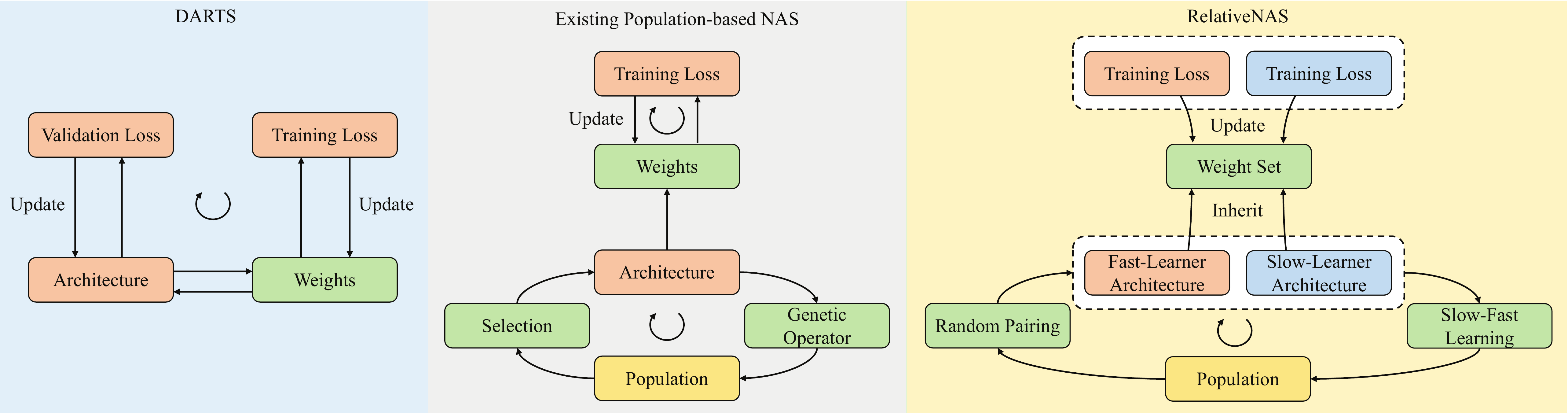}
     \caption{Illustration of the three general frameworks of different NAS methods. 
                LEFT: DARTS updates the solo candidate architecture and its weights simultaneously by gradients generated by validation loss and training loss respectively.
                MIDDLE: existing population-based NAS updates a population of architectures by stochastic crossover/mutation.
                In order to obtain the ranking of the architectures in each generation, the newly generated architectures need to be trained on the training set for a certain number of epochs.
                RIGHT: RelativeNAS updates a population of architectures by the proposed \emph{slow-fast learning paradigm} (instead of gradient or crossover/mutation).
                For each pair of architectures, slow-learner is distinguished from fast-learner by their relative performances, where for such low-fidelity performance estimation, each architecture only needs to inherit its weights from a weight set, and is trained by only one epoch to update the weights.
%
%
                } \label{fig:framework_comp}
\end{figure*}

\IEEEPARstart{D}{eep} Convolutional Neural Networks (CNNs) have achieved remarkable results in various computer vision tasks (\emph{e.g.}, image classification~\cite{krizhevsky2012imagenet,kong2020constructing,muxconv}, object detection~\cite{ren2015faster}, and semantic segmentation~\cite{long2015fully, yu2018bisenet}), and a number of state-of-the-art networks have been designed by experts since 2012~\cite{szegedy2015going,he2016deep,huang2017densely}.
Since the manual designs of CNNs heavily rely on expert knowledge and experience, it is usually time-consuming and error-prone.
To this end, researchers have turned to automatic generation of high-performance network architectures for any given tasks, a.k.a. Neural Architecture Search (NAS)~\cite{elsken2019neural}.
Without loss of generality, the problem of NAS for a target dataset $\mathcal{D}=\{\mathcal{D}_{trn},\mathcal{D}_{vld},\mathcal{D}_{tst}\}$ and a search space $\mathcal{A}$ can be formulated as~\cite{lu2019multi}:
\begin{eqnarray}\label{eq:NAS}
\text{minimize}&&\mathcal{L}_{vld}(\boldsymbol{\alpha},\boldsymbol{\omega}^*)\\
\text{subject to}&&\boldsymbol{\omega}^*\in \arg\min\limits_{\boldsymbol{\omega} }{\mathcal{L}_{trn}(\boldsymbol{\alpha}, \boldsymbol{\omega})} , \nonumber
\end{eqnarray}
where ${\boldsymbol{\alpha} \in \mathcal{A}}$ defines the model architecture, $\boldsymbol{\omega}$ defines the associated weights, and $\boldsymbol{\omega}^*$ defines the corresponding optimal weights.
Besides, $\mathcal{D}_{trn}$, $\mathcal{D}_{vld}$, and $\mathcal{D}_{tst}$ are the training data, validation data, and test data, respectively.
$\mathcal{L}_{trn}$ and $\mathcal{L}_{vld}$ denote the loss on the training data and validation data respectively.

Despite the promising performance of recent NAS methods on image classification~\cite{zhong2020blockqnn,lu2021nat}, object detection~\cite{ghiasi2019fpn, xu2019auto}, semantic segmentation~\cite{liu2019auto, chen2019fasterseg}, and designing generative adversarial networks~\cite{gong2019autogan}, there are two major challenges: 
(1) NAS is treated as a black-box optimization problem due to the lack of a priori knowledge adopted with the exact functional relationship between architectures and their performances;
(2) NAS suffers from high computation costs due to a large number of performance estimations of candidate architectures.

Among various NAS methods, the differentiable NAS (\emph{i.e.} DARTS)~\cite{liu2018darts} and population-based NAS~\cite{real2018regularized} are among the most popular ones due to their unique merits for tackling each challenge: 
DARTS mainly benefits from the merit of high search efficiency due to relaxing the search space to be continuous; 
population-based NAS mainly benefits from the merit of diversified candidate architectures in the population and involving genetic operators (\emph{e.g.}, crossover/mutation) to drive the search process.
Nevertheless, they also suffer from some deficiencies: since DARTS jointly trains a supernet and search for an optimal solution merely by gradient, it suffers from low robustness in terms of flexibility and versatility; population-based NAS mainly relies on stochastic crossover/mutation for search, and it usually requires a large amount of computation cost for performance evaluations.

One research question is: \emph{can we benefit from the merits of both differentiable NAS and population-based NAS while overcoming their deficiencies? }
To answer it, this work proposes a novel RelativeNAS method, as shown in Fig.~\ref{fig:framework_comp} (right).

\begin{figure}[!htb]
     \centering
     \includegraphics[width=0.9\linewidth]{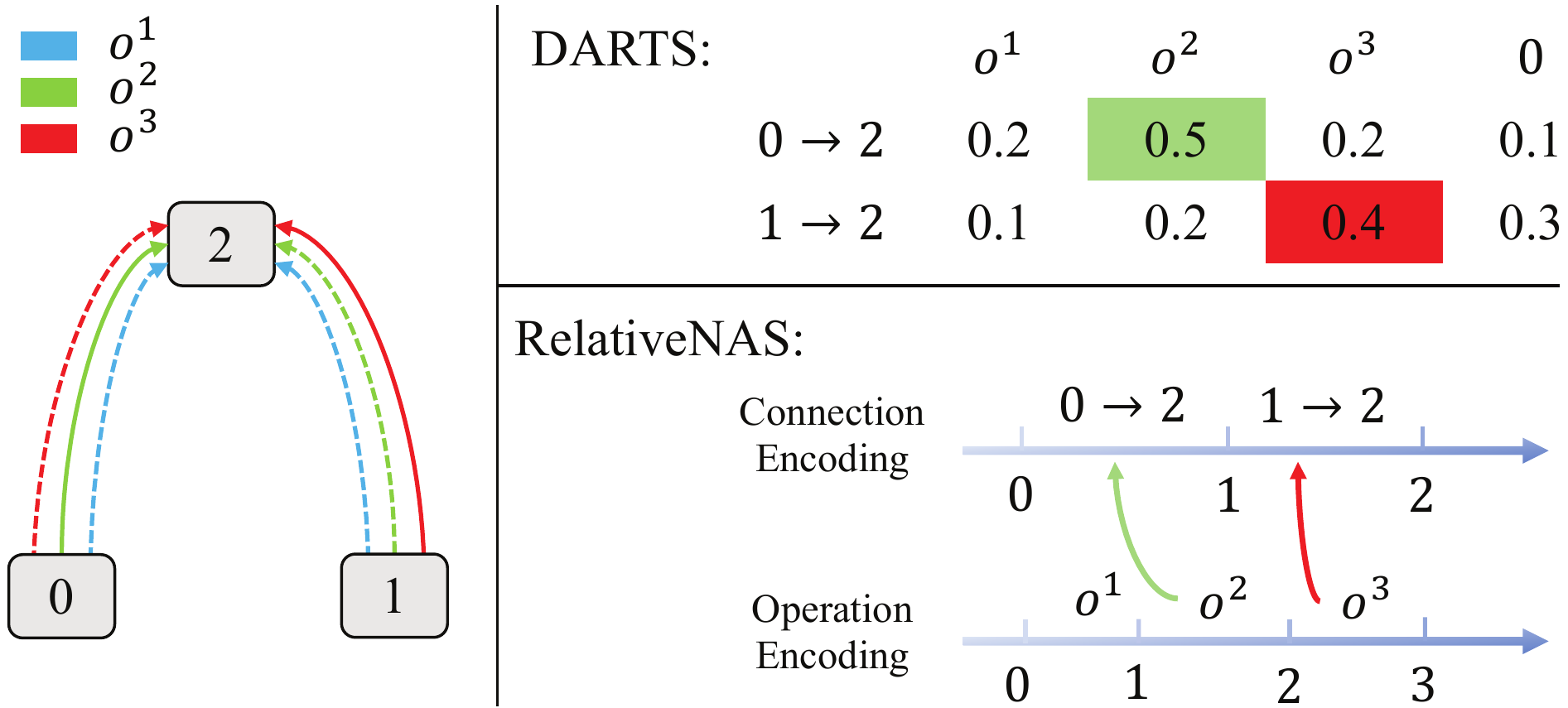}
     \caption{Examples of the encoding schemes of DARTS~\cite{liu2018darts} and RelativeNAS.
                LEFT: an architecture to be encoded.
                Boxes, dashed lines, and solid lines represent different nodes, the candidate operations, and chosen operations respectively.
                Node $2$ needs to choose two operations (\emph{i.e.} two out of the six lines). 
                RIGHT: the two different continuous encoding schemes.
                $a\rightarrow b$ represents the connection between Node $a$ and Node $b$ while $o^1$, $o^2$, and $o^3$ denote the candidate operations.
                DARTS encodes every operation by its weight and uses $0$, \emph{i.e.} $zero$ operation, to denote that there is no connection between two nodes; each real number means the probability/weight of choosing the corresponding operation.
                RelativeNAS uses four continuous variables, including two connections and their corresponding operations, to encode the architecture; the real number falling into each interval means that the corresponding connection exists or the corresponding operation is chosen.
                Note that, since each node in RelativeNAS must connect to two predecessor nodes, $zero$ operation is not considered in the search space. 
              }\label{fig:encoding_comp}
\end{figure}

Particularly, this work proposes a novel continuous encoding scheme for cell-based search space by considering connections between pairwise nodes and the corresponding operations, inspired by~\cite{liu2018darts}. 
However, in contrast to the encoding method as given in~\cite{liu2018darts}, the proposed one has no requirement of differentiability, neither does it consider the probability/weight of choosing an operation; instead, it directly encodes the operations between pairwise nodes into real values in a naive manner (as shown in Fig.~\ref{fig:encoding_comp}).
The main advantages of the proposed continuous encoding method are: (1) it provides more flexibility and versatility; (2) the enlarged search space encourages the search for diverse architectures when applied to population-based NAS~\cite{stanley2019designing}.

With the proposed continuous encoding scheme, this work further proposes a slow-fast learning paradigm for efficient search in the encoded space, inspired by~\cite{feichtenhofer2019slowfast}.
In this paradigm, a population of architecture vectors is iteratively paired and updated.
Specifically, in each iteration of the proposed slow-fast learning, the architecture vectors are randomly paired; for each pair of architecture vectors, the one with worse performance (denoted as slow-learner) is updated by learning from the one with better performance (denoted as fast-learner).
In contrast to the population-based NAS methods such as large-scale evolution~\cite{real2017large} and AmoebaNet~\cite{real2018regularized}, the proposed slow-fast learning paradigm does not involve any genetic operator (\emph{e.g.}, crossover/mutation), but essentially, the architecture vectors are updated using a pseudo-gradient mechanism which aims to learn the joint distribution between each pair of slow-learner and fast-learner implicitly.
Specifically, the pseudo-gradient is determined by the pairwise learning between the fast-learner and slow-learner.
The main advantages of the proposed slow-fast learning paradigm are: (1) it provides a scheme to perform NAS in generic continuous search space without considering its specific properties (\emph{e.g.} differentiability); (2) it suggests a way to learn joint distributions of multiple architectures.

To improve the computation efficiency of RelativeNAS, this work further adopts a weight set as a knowledge base to estimate the performances when comparing the architectures in each pair, where the weight set is an operation collection of all the candidate architectures, as well as a gathering of the promising knowledge in the population.
Since the slow-fast learning only requires low-fidelity performance estimation when distinguishing slow-learner and fast-learner in each pair, a newly discovered network is trained for only one epoch to obtain the estimated performance, thus saving substantial computation cost for performance evaluations.
It is worth noting that the weight set is not directly trained but updated by the trained weights of each paired networks in an online manner.
The contributions of this paper can be summarized as follows:
\begin{enumerate}
    \item 
    We propose a novel continuous encoding scheme and a slow-fast learning paradigm for NAS.
    Specifically, a population of architecture vectors is iteratively updated by pseudo-gradients,  towards optima.
    \item 
    We propose a novel performance estimation strategy to expedite the evaluation of architectures, where the weight set is not directly trained but updated using other trained weights. 
    It couples weight set updating and searching into a single integrated online process.
    \item 
    The proposed RelativeNAS significantly reduces the general computation cost of NAS, spending only nine hours with a single 1080Ti GPU to obtain the discovered architecture on CIFAR-$10$.
    Furthermore, we show that the searched architecture achieves state-of-the-art performance (97.66\% accuracy) on CIFAR-$10$.
    \item 
    We transfer the searched architecture to multiple intra- and inter-domain tasks, including image classification, object detection, image segmentation, and keypoint detection. 
    The consistently better performances across these tasks demonstrate that the improvement on CIFAR-$10$ is an actual advancement from the architecture as obtained by RelativeNAS.
\end{enumerate}

The rest of the paper is organized as follows.
The background knowledge, including some related works and the motivation of this work, is given in Section~\ref{sec:background}.
The details of the proposed RelativeNAS are elaborated in Section~\ref{sec:method}.
Experimental studies are presented in Section~\ref{sec:result}.
Finally, the conclusions are drawn in Section~\ref{sec:conclusion}.

\section{Background}\label{sec:background}

In this section, we first present some related works, including those in differentiable NAS and population-based NAS.
Then, we briefly summarize the motivation of this work.

\subsection{Differentiable Neural Architecture Search}

Differentiable NAS is motivated to relax the architecture into continuous and then use gradient-based approaches for optimization.
DARTS~\cite{liu2018darts} is a pioneering work in differentiable NAS, where three components have significantly improved the computation efficiency: the cell-based search space, the continuous relaxation approach, and the approximation technique.
Specifically, the cell-based search space modularizes the entire CNN into a stack of several cells to reduce the number of parameters to be optimized; the continuous relaxation schema transforms the choices of discrete operations into a differentiable learning objective for the joint optimization of the architecture and its weights; moreover, the approximation technique approximates the architecture gradient for reducing the expensive inner optimization.

Following DARTS, SNAS~\cite{xie2018snas} has proposed to optimize the architecture distribution for the operations during the back-propagation.
Specifically, the search space is differentiable by relaxing a set of one-hot random variables, which is used to select the corresponding operations in the graph.
Since SNAS has used \emph{search gradient} as the reward instead of \emph{training loss} for reinforcement learning, the objective is not changed but the process is more efficient.
ProxylessNAS~\cite{cai2018proxylessnas} has used the binarized architecture parameters to reduce the GPU memory.
Besides, it makes latency differentiable and adds the expected latency into the loss function as a regularization term. 

However, a recent work~\cite{liang2019darts+} has discovered that the bi-level optimization of weights and architecture in DARTS will collapse when the search epochs become larger.
Since the number of weight parameters is larger than the number of architecture parameters, weight optimization will restraint the architecture optimization, and more \emph{identity} operations are chosen to deteriorate the performance.
Essentially, it is mainly attributed to the lack of diversity when the solo candidate architecture in DARTS is optimized by the gradient.

\begin{algorithm}[t]
\caption{RelativeNAS Framework}
\label{al:framework}
\begin{algorithmic}[1]
\REQUIRE Training set $\mathcal{L}_{trn}$, validation set $\mathcal{L}_{vld}$, population size $N$, generation number $G$

\STATE Initialize a population $\{\boldsymbol{\alpha}^0\}_{n=1}^N$ and a weight set $\Omega$;
\FOR{$g = 1,2,\dots, G$}
 \STATE $\{\boldsymbol{\alpha^g}\}_{n=1}^N$ is randomly divided into $N/2$ pairs;
 \FOR{$p = 1,2,\dots, N/2$}
  \STATE The $p$-th pair of encoded vectors $\{\boldsymbol{\alpha}_{p,j}^g\}_{j=1,2}$ are decoded into networks $\{C_{\boldsymbol{\alpha}_{p,j}^g}\}_{j=1,2}$;
  \STATE $\{C_{\boldsymbol{\alpha}_{p,j}^g}\}_{j=1,2}$ inherit weights $\{\boldsymbol{\omega}_{\boldsymbol{\alpha}^g_{p,j}}\}_{j=1,2}$ from weight set $\Omega$;
  
   /* train each network with one epoch */
  \FOR{$j = 1,2$}
  \STATE  $\boldsymbol{\omega}'_{\boldsymbol{\alpha}^g_{p,j}}=step(\boldsymbol{\omega}_{\boldsymbol{\alpha}^g_{p,j}}|C_{\boldsymbol{\alpha}^g_{p,j}})$;
  \STATE $\mathcal{L}_{\boldsymbol{\alpha}^g_{p,j}} = \mathcal{L}_{vld}(C_{\boldsymbol{\alpha}^g_{p,j}},\boldsymbol{\omega}'_{\boldsymbol{\alpha}^g_{p,j}})$;

 \ENDFOR

 /* distinguish slow-learner and fast-learner */
 \STATE $\boldsymbol{\alpha}^g_{p,f} = \arg\min\limits_{\boldsymbol{\alpha}}(\mathcal{L}_{\boldsymbol{\alpha}^g_{p,1}},\mathcal{L}_{\boldsymbol{\alpha}^g_{p,2}})$;
  \STATE $\boldsymbol{\alpha}^g_{p,s} = \arg\max\limits_{\boldsymbol{\alpha}}(\mathcal{L}_{\boldsymbol{\alpha}^g_{p,1}},\mathcal{L}_{\boldsymbol{\alpha}^g_{p,2}})$;
  
 /* update weight set $\Omega$ */
 \STATE $\Omega$ is updated with (\ref{eq:update_weight_set});

/* slow-learner learns from fast-learner */
 \STATE $\boldsymbol{\alpha}^g_{p,s}$ is updated with (\ref{eq:arch_update_pair});

 \ENDFOR
\ENDFOR

\end{algorithmic}
\end{algorithm}

\begin{figure*}[!htb]
    \centering
    \includegraphics[width=0.8\linewidth]{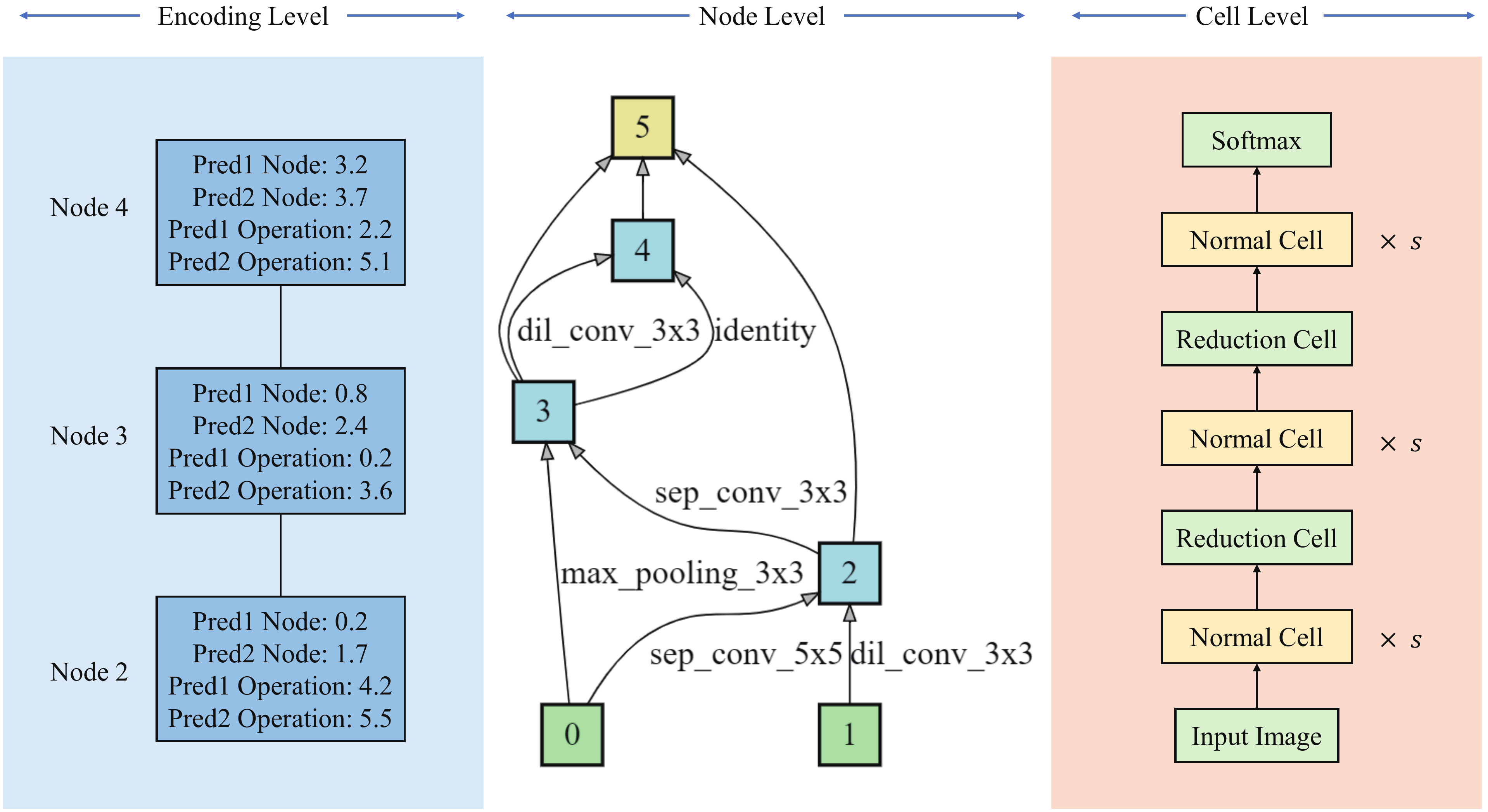}
    \caption{An example of an encoded vector which maps to the intermediate nodes of the cell-based structure.
    LEFT: An encoded vector with three blocks. 
    MIDDLE: the corresponding architecture by the encoded vector with three intermediate nodes.
    RIGHT: the overall cell-based architecture.}\label{fig:encoding_exampe}
\end{figure*}

\subsection{Population-Based Neural Architecture Search}
As indicated by the term itself, population-based NAS maintains a population where each individual inside it represents a candidate architecture.
Cooperation among candidate architectures modifies their attributes, and competition for removing the worse and retaining the better pushes the population towards optima. 
Neuroevolution is a traditional approach that evolves the neural network topologies and their weights simultaneously~\cite{yao1999evolving,stanley2002evolving}.
Along with the deeper layers of the neural networks and increasing parameters, evolving the weights becomes prohibitively time-consuming.
Hence, population-based NAS merely evolves the neural network topologies but train the neural networks using the back-propagation method.

Most existing population-based NAS methods have adopted the genetic algorithm or genetic programming~\cite{suganuma2017genetic,lu2019nsga,nsganetv2} to mimic the process of natural evolution, which requires the elaborate designs of stochastic crossover/mutation operators. 
The research in~\cite{real2017large} is a pioneering work in population-based NAS, where $11$ mutation operators are designed to modify the attributes of the networks, including altering learning rate, resetting weights, inserting convolution, removing convolution, etc.
This research has indicated good feasibility in designing different genetic operators, showing that a neural network could be evolved starting from a very simple form and growing into complex architecture.
Afterwards, AmoebaNet~\cite{real2018regularized} discovered an architecture that surpassed the human-craft architectures by using NAS for the first time.
In AmoebaNet, a macro architecture is predefined to comprise a number of identical \emph{cells}, such that the search space is reduced to the cell architecture instead of the entire one. 
On top of such a cell-based framework, two different kinds of mutation operators are designed to change the operation types and the connections among different nodes.

Despite the promising performance of population-based NAS, the existing methods mainly suffer from low computation efficiency.
This can be attributed to two factors.
First, the stochastic crossover/mutation, as commonly adopted in existing population-based NAS methods, can be inefficient in generating high-performance candidate architectures.
Second, the performance evaluations of newly generated candidate architectures can be computationally expensive.

Several methods have been proposed in order to improve the efficiency of the population-based NAS.
Lower fidelity estimates are common strategies for speeding up NAS, where the performance is estimated on the basis of lower fidelities of the actual performance, including a reduced number of training epochs, models with smaller sizes, or training on a subset of the datasets.
For example, AmoebaNet~\cite{real2018regularized} uses each model trained for 25 epochs with a fewer number of cells. 
Another strategy to speed up the estimation of architectures is the one-shot approach~\cite{brock2017smash,bender2018understanding}, where only the weights of a single supernet are trained.
The searchable architectures can then be evaluated using the weights inherited from the supernet without further training.


\subsection{Motivation}
While differential NAS has opened a new dimension in the literature, its development is still in its infancy. 
One pivotal issue is how to make the best of the continuously encoded search space.
Due to the limitations of back-propagation, searching over one solo architecture by gradient can be ineffective due to the lack of proper diversity.
By contrast, population-based NAS is intrinsically advanced in maintaining diversity when searching over multiple candidate architectures, but its search efficiency is poor due to the stochastic crossover/mutation and a large number of performance evaluations.
Therefore, this work is essentially motivated to design an effective and efficient NAS method, which benefits from the merits of both differentiable NAS and population-based NAS while overcoming their deficiencies.

\section{Methodology} \label{sec:method}

We present the pseudo-code of the proposed slow-fast learning paradigm in Algorithm~\ref{al:framework}.
The proposed RelativeNAS mainly involves three pivotal components: encoding/decoding of the search space (Line $1$ and Line $5$), performance estimation of the candidate architectures (Line $6$ to Line $13$), and slow-fast learning among the architecture vectors (Line $14$).
The computational complexity of performance estimation strategy mainly depends on the task (\emph{i.e.}, classification, detection, \emph{etc}.) and the scale of dataset.
Apart from the performance estimation strategy, another major computational complexity in RelativeNAS is associated with the update of the slow-learner, which is an inevitable operation for population-based NAS.
Accordingly, the computational complexity of updating the population is $\mathcal{O}(N*|\boldsymbol{\alpha}|)$, where $N$ is the population size and $|\boldsymbol{\alpha}|$ is the dimensionality of the searched architecture vector.
In the following subsections, we will elaborate the three components respectively.

\subsection{Search Space}\label{ssec:search_space}

\begin{figure*}[!htb]
     \centering
     \includegraphics[width=1\linewidth]{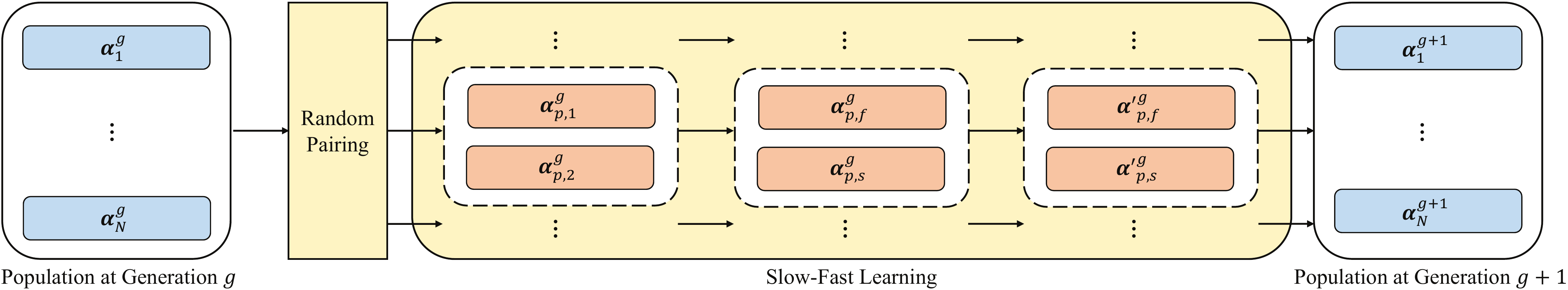} 
     \caption{Illustration of the slow-fast learning process at generation $g$.
        A population of $N$ architecture vectors is randomly divided into $N/2$ pairs.
        For each pair $p$, slow-learner $\boldsymbol{\alpha}_{p,s}^g$ updates its vector by learning from a fast-learner $\boldsymbol{\alpha}_{p,f}^g$ using (\ref{eq:general_upadate}) and (\ref{eq:arch_update_pair}), while fast-learner $\boldsymbol{\alpha}_{p,f}^g$ itself remains unchanged.
       After the slow-fast learning process, all fast-learners and updated slow-learners are re-merged to become the new population of the next generation $g+1$.
       }\label{fig:arch_update}
\end{figure*}

To limit the scale of search space, this work adopts the cell-based architecture inspired by~\cite{liu2018darts}.
Specifically, There are two types of cells: the normal cell $\mathcal{G}^n$ and the reduction cell $\mathcal{G}^r$.
The only difference between the two types of cells is the output feature size.
The normal cell does not change the size of the feature, while the reduction cell is served as down-sampling to reduce the size of the feature by stride operation.
The internal structure of the two types of cells is represented by a directed acyclic graph.
As shown in Fig.~\ref{fig:encoding_exampe} (middle), there are two input nodes from the two previous cells.
Every intermediate node contains two predecessor nodes and applies operations on them. 
Consequently, the edge in the graph is denoted as the possible operation and information flow between different nodes.
Edges are only allowed to point from low indexed nodes to higher ones.
On the other hand, a cell only contains one output and all the intermediate nodes are concatenated to the output node.
Besides, the reduction cell is connected after $s$ normal cells as shown in Fig.~\ref{fig:encoding_exampe}  (right).
In detail, the cell-based search space can be formulated as:
 \begin{align*}
 	\mathcal{G}_c =& \{\mathcal{G}_{1}^n,...,\mathcal{G}_{s}^n, \mathcal{G}_{s+1}^r,\\
 	              & \mathcal{G}_{s+2}^n,...,\mathcal{G}_{2s+1}^n,\mathcal{G}_{2s+2}^r,\\
 	              & \mathcal{G}_{2s+3}^n,...,\mathcal{G}_{3s+2}^n\}.
 \end{align*}

Since there are $3s+2$ different cells in total, the search space can be too large for efficient NAS.
To address this issue, all the normal cells are identified, so are the reduction cells.
Therefore, the cell-based search space is constrained to a normal cell $\mathcal{G}^n$ and a reduction cell $\mathcal{G}^r$ as below:
 \begin{align*}
 	\mathcal{G}_c  = \{\mathcal{G}^n, \mathcal{G}^r\}.
 \end{align*}
Specifically, such a cell-based architecture has the following advantages.
First, it maintains diversity inside the cells. 
While all of them share the same macro architecture, the different connections among nodes and operations inside the two types of cells diversify the structure of the network.
Second, under the cell-based design, networks can achieve promising performance and high transferability for different tasks by adjusting the total number of cells in the final architecture. 

\begin{table}[!t]

\renewcommand{\arraystretch}{1.3}
\caption{Illustration of the encoding scheme in RelativeNAS. 
An operation is determined by the operation type and its kernel size.
Each node or operation corresponds to a unique range in real number space.
}
\label{code_value}
\centering
\begin{tabular}{cc|ccc}
\toprule 
\multicolumn{2}{c|}{Node} & \multicolumn{3}{c}{Operation}\\
\midrule
Range & Index & Range & Type & Kernel Size \\
\hline
$[0,1)$ & $0$ & $[0,1)$ & Max Pooling & $3$\\
\hline
$[1,2)$ & $1$ & $[1,2)$ & Avg Pooling & $3$\\
\hline
$[2,3)$ & $2$ & $[2,3)$ & Identity & $0$\\
\hline
$[3,4)$ & $3$ & $[3,4)$ & Sep Conv & $3$\\
\hline
$[4,5)$ & $4$ & $[4,5)$ & Sep Conv & $5$\\
\hline
$[5,6)$ & $5$ & $[5,6)$ & Dil Conv & $3$\\
\hline
$[6,7)$ & $6$ & $[6,7)$ & Dil Conv & $5$\\
\bottomrule
\end{tabular}
\end{table}

To transform a directed acyclic graph into uniform representation in the continuous space, this work proposes a novel continuous encoding scheme, as illustrated in Table~\ref{code_value}. 
Since the input nodes and the output node are fixed, this work only needs to encode every intermediate node along with its two predecessor nodes and the corresponding operations.
To be more specific, this work encodes the node and the operation separately.
Each node or operation is represented by a real value interval.
Each interval is left-closed and all intervals added together are continuous. 
To guarantee uniqueness, there is no overlap among different intervals of the nodes or operations.
Without bias, each interval of a node or an operation has the same length.

Specifically, there are seven different operations, including $3\times3$ max pooling, $3\times3$ average pooling, two depth-wise separable convolutions~\cite{chollet2017xception} (Sep Conv $3\times3$, $5\times5$), two dilated separable convolutions (Dil Conv $3\times3$ and $5\times5$), and identity.
Unless specified, each convolutional layer in the network is fronted by ReLU activation and followed by batch normalization~\cite{ioffe2015batch},  and each separable convolution is applied twice. 

With the proposed encoding scheme, the cell-based search space $\mathcal{G}_c$ is encoded into a new one $\mathcal{A}$, \emph{i.e.}the encoded search space.
In this way, every architecture vector $\boldsymbol{\alpha} \in \mathcal{A} $ can be decoded into its corresponding cell architecture $C_{\boldsymbol{\alpha}} \subset \mathcal{G}_c$ by mapping the vector into the connections and operations of the intermediate nodes in the cell. 

An illustrative example of the above encoding process is given in Fig.~\ref{fig:encoding_exampe}.
This work uses a list of blocks to represent an architecture vector as shown in Fig.~\ref{fig:encoding_exampe} (left).
Each block represents an intermediate node in the cell and needs to be specified by four variables, including Pre1 Node (the first predecessor node) and Pre2 Node (the second predecessor node), and their corresponding operations.
Fig.~\ref{fig:encoding_exampe} (middle) shows a cell architecture decoded from Fig.~\ref{fig:encoding_exampe} (left) using the mapping rules in Table~\ref{code_value}.

\subsection{Slow-Fast Learning}\label{arch_update}

The general target of NAS is to search for an architecture vector $\boldsymbol{\alpha}^\ast \in \mathcal{A}$ such that the decoded architecture $C_{\boldsymbol{\alpha}^*}$ minimizes validation loss $\mathcal{L}_{vld}(C_{\boldsymbol{\alpha}^*},\boldsymbol{\omega}^*_{\boldsymbol{\alpha}^*})$, where the weights $\boldsymbol{\omega}^*_{\boldsymbol{\alpha}^*}$ associated with the architecture are obtained by minimizing the training loss $\mathcal{L}_{trn}(C_{\boldsymbol{\alpha}^*},\boldsymbol{\omega}_{\boldsymbol{\alpha}^*})$. 
To this end, the slow-fast learning paradigm in the proposed RelativeNAS is essentially an optimizer to approximate $\boldsymbol{\alpha}^*$.

Specifically, given an architecture vector $\boldsymbol{\alpha}^{g}$ obtained at the $g$-th generation\footnote{To distinguish the \emph{iteration} in network training, this work uses the term \emph{generation} to denote the iteration in slow-fast learning}  of slow-fast learning, it is iteratively updated by:
\begin{equation}
\boldsymbol{\alpha}^{g+1} = \boldsymbol{\alpha}^{g} + \Delta \boldsymbol{\alpha}^g, \label{eq:general_upadate}
\end{equation} 
where $\mathcal{L}_{vld}(C_{\boldsymbol{\alpha}^{g+1}},\boldsymbol{\omega}^*_{\boldsymbol{\alpha}^{g+1}}) < \mathcal{L}_{vld}(C_{\boldsymbol{\alpha}^{g}},\boldsymbol{\omega}^*_{\boldsymbol{\alpha}^{g}})$ holds, such that: 
\begin{equation}
{\lim_{g \to +\infty}} \boldsymbol{\alpha}^g = \boldsymbol{\alpha}^*.
\end{equation}
To efficiently generate the pseudo gradient $\Delta\boldsymbol{\alpha}^{g}$, this work proposes to use a population of $N$ architecture vectors $\{\boldsymbol{\alpha}\}_{n=1}^N$. 
As illustrated in Fig.~\ref{fig:arch_update}, at each generation, the population is randomly divided into $N/2$ pairs.
Then, for each pair $p$, a fast-learner $\boldsymbol{\alpha}_{p,f}^g$ and a slow-learner $\boldsymbol{\alpha}_{p,s}^g$ are specified by the paritial ordering of validation loss values, where the one having smaller loss is the fast-learner and the other is the slow-learner (\emph{i.e.} $\mathcal{L}_{vld}(C_{\boldsymbol{\alpha}^{g}_{p,f}},\boldsymbol{\omega}^*_{\boldsymbol{\alpha}^{g}_{p,f}}) <  \mathcal{L}_{vld}(C_{\boldsymbol{\alpha}^{g}_{p,s}},\boldsymbol{\omega}^*_{\boldsymbol{\alpha}^{g}_{p,s}})$).
Then, $\boldsymbol{\alpha}_{p,s}^g$ is updated by learning from $\boldsymbol{\alpha}_{p,f}^g$ with:
\begin{equation}\label{eq:arch_update_pair}
\Delta \boldsymbol{\alpha}_{p,s}^{g} = \lambda_1 (\boldsymbol{\alpha}_{p,f}^g-\boldsymbol{\alpha}_{p,s}^{g}) + \lambda_2\Delta\boldsymbol{\alpha}_{p,s}^{g-1}, 
\end{equation}
where $\lambda_1, \lambda_2 \in [0,1]$ are randomly generated values by uniform distribution. 
Specifically, $\lambda_1$ determines the step size that $\boldsymbol{\alpha}^g_{p,s}$ learns from $\boldsymbol{\alpha}^g_{p,f}$, and $\lambda_2$ determines impact of the momentum term $\Delta\boldsymbol{\alpha}^{g-1}_{p,s}$.
Such a pseudo gradient is inspired by the second derivatives in the gradient descent of the back propagation~\cite{rumelhart1986learning}.
Thanks to such a pseudo-gradient based mechanism, the proposed RelativeNAS is applicable not only to the search space in this work, but also to any other generic continuously encoded search space.

Eventually, all fast-learners and updated slow-learners are re-merged to become the new population of the next generation $g+1$.
By such an iterative process of slow-fast learning, each architecture vector in the population is expected to move towards optima by learning from those converging faster than them.

\begin{figure}[!htb]
     \centering
     \includegraphics[width=0.98\linewidth]{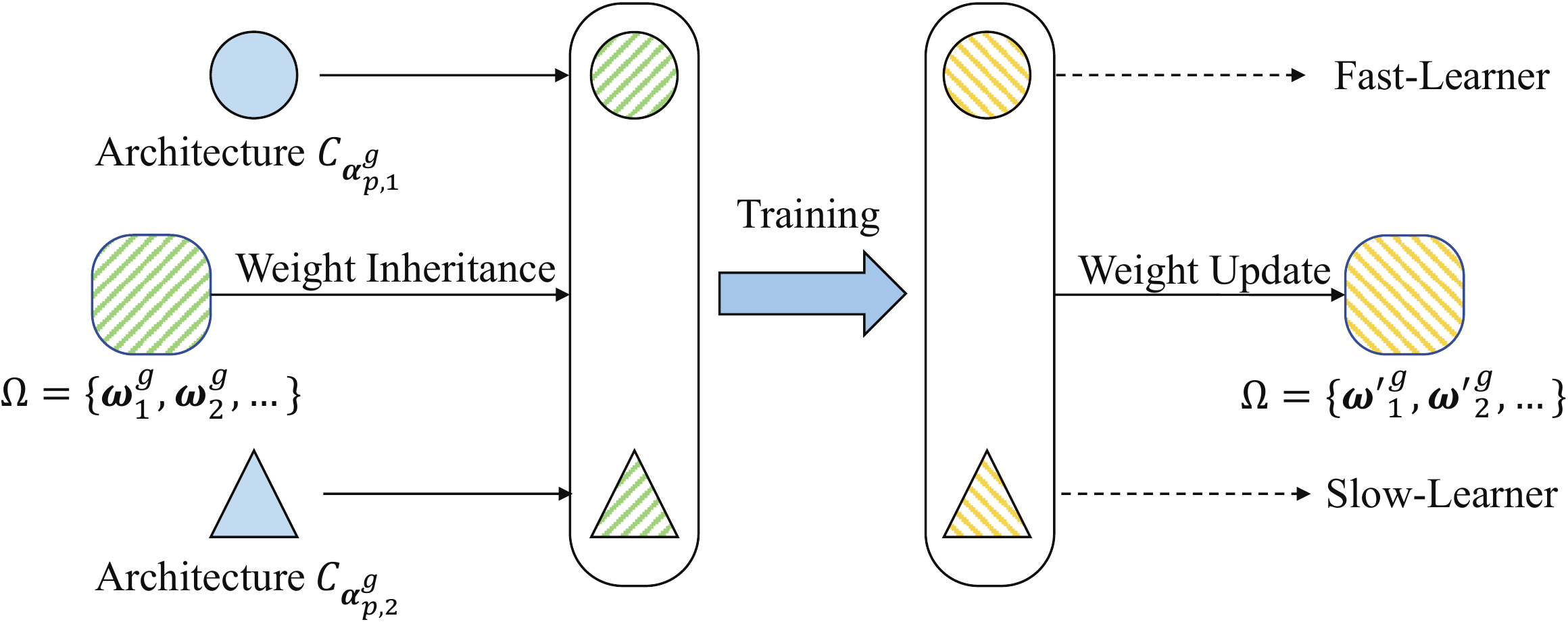}
     \caption{Illustration of the architecture evaluation process.
        Given two paired architectures $C_{\boldsymbol{\alpha}^g_{p,1}}$ and $C_{\boldsymbol{\alpha}^g_{p,2}}$, they inherit weights from the weight set $\Omega$ firstly.
        After that, we train each network only for one epoch and differ them to fast-learner and slow-learner.
        Finally, the weight set is updated by all the trained weights $\boldsymbol{\omega}'^g$. 
        }\label{fig:arch_eval}
\end{figure}

\subsection{Performance Estimation}\label{arch_eval}
As described above, the proposed RelativeNAS needs to evaluate the performances of candidate architectures thus decoded from the architecture vectors in the population of each generation, such that for each pair of architecture vectors, fast-learner can be distinguished from slow-learner by their validation losses. 
Ideally, the performance of a candidate architecture $C_{\boldsymbol{\alpha}^g}$ can by evaluated by solving the following optimization problem:  
\begin{align}
	\boldsymbol{\omega}_{\boldsymbol{\alpha}^g}^* &= optimize(\boldsymbol{\omega}_{\boldsymbol{\alpha}^g}| C_{\boldsymbol{\alpha}^g}) \nonumber\\
	&= step(step(...step(\boldsymbol{\omega}_{\boldsymbol{\alpha}^g}|C_{\boldsymbol{\alpha}^g})...|C_{\boldsymbol{\alpha}^g})| C_{\boldsymbol{\alpha}^g}), \label{eq:sgd}
\end{align}
where $\boldsymbol{\omega}_{\boldsymbol{\alpha}^g}^*$ is the optimal weights of the candidate architecture,  and function $step(\cdot)$ is one step\footnote{In this work, one \emph{step} is specified as one \emph{epoch} in the training process.} of the iterative optimization procedure to update the weights of the neural network.
In practice, however, solving such an optimization (\emph{i.e.} training the candidate architecture) can be quite computationally expensive, especially when there are a large number of candidate architectures obtained during the iterative slow-fast learning process.
Hence, to reduce the computation cost of performance evaluations in RelativeNAS, this work proposes a performance estimation strategy.

\begin{figure*}[!htb]
     \centering
     \includegraphics[width=1\linewidth]{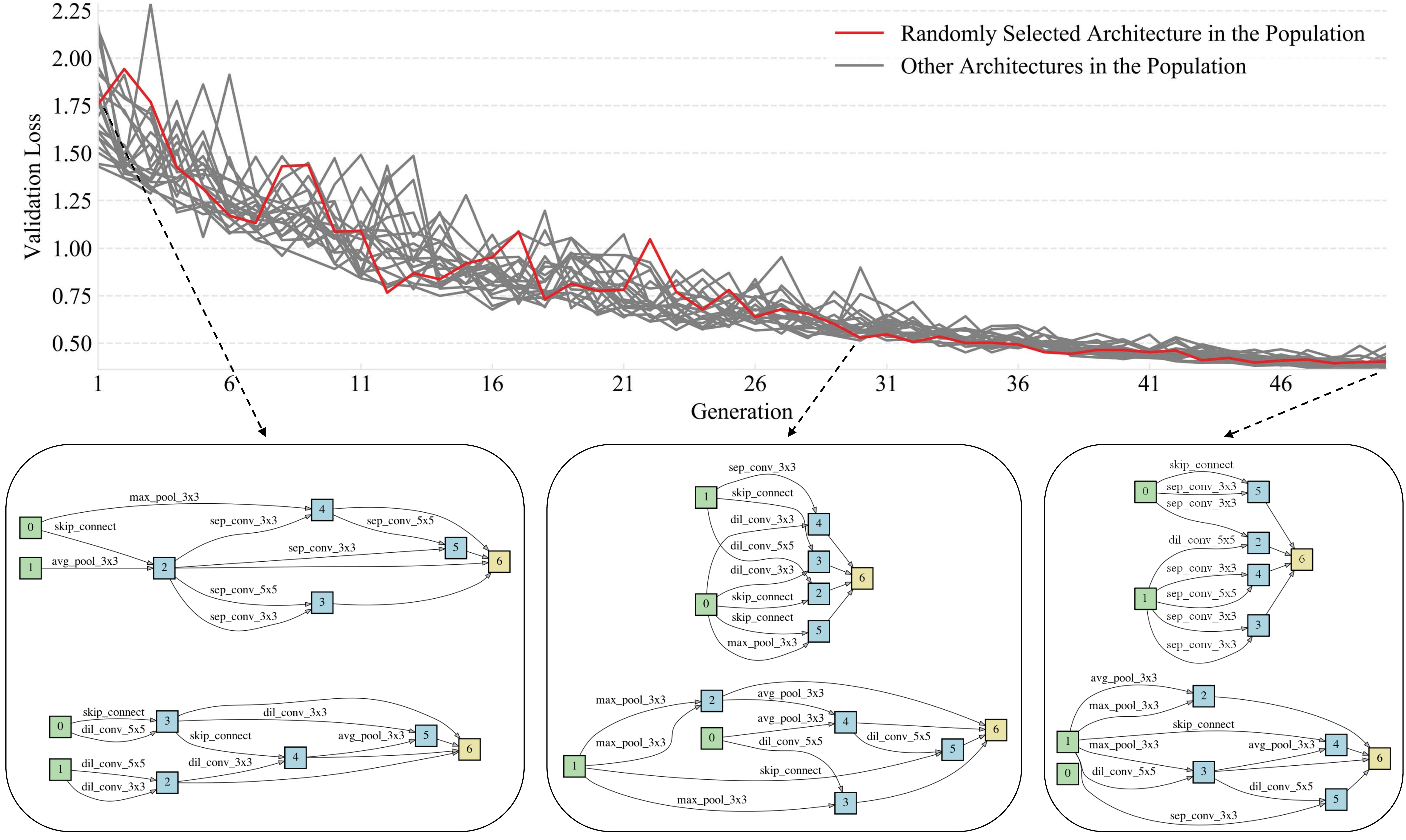}
     \caption{Searching process on CIFAR-$10$. 
     TOP: trajectories of validation losses using our performance estimation method for all candidate architectures decoded from the architecture vectors in the population. The \textcolor{red}{red line} indicates a randomly selected architecture in the population.
     BOTTOM: architectures of a randomly selected candidate architecture in the population at generation $1$, $30$, and $50$ respectively.
     Each box contains a normal cell (top) and reduction cell (bottom).}\label{fig:pop_data}
\end{figure*}

In contrast to existing differentiable NAS methods (\emph{e.g.} DARTS~\cite{liu2018darts}), the validation losses in RelativeNAS are not directly involved in updating the candidate architectures; instead, they are merely used to determine the \emph{partial orders} among each pair of candidate architectures (\emph{i.e.} to distinguish fast-learner and slow-learner).
Therefore, in RelativeNAS, it is intuitively feasible to use performance estimations (instead of exact performance evaluations) to obtain the approximate validation losses of the candidate architectures.

Specifically, the proposed performance estimation strategy first randomly initializes a weight set $\Omega$ to contain the weights of all possible operations in $\mathcal{G}_c$.
During the search process, given the $p$-th pair of candidate architecture $\{C_{\boldsymbol{\alpha}^g_{p,j}}\}_{j=1,2}$ at generation $g$, they inherit the corresponding weights $\boldsymbol{\omega}_{\boldsymbol{\alpha}^g_{p,1}}$ and $\boldsymbol{\omega}_{\boldsymbol{\alpha}^g_{p,2}}$ from $\Omega$ according the their own operations respectively.
With the inherited weights as warm-up, the weights of $\{C_{\boldsymbol{\alpha}^g_{p,j}}\}_{j=1,2}$ only need to be updated on training set $\mathcal{D}_{trn}$ by one step of optimization:
\begin{gather*}\label{weight_update}
  \boldsymbol{\omega}'_{\boldsymbol{\alpha}^g_{p,j}}=step(\boldsymbol{\omega}_{\boldsymbol{\alpha}^g_{p,j}}|C_{\boldsymbol{\alpha}^g_{p,j}}).
 \end{gather*}
Then, the updated weights $\{\boldsymbol{\omega}'_{\boldsymbol{\alpha}^g_{p,j}}\}_{j=1,2}$ are used to estimate the validation losses $\{\mathcal{L}_{vld}(C_{\boldsymbol{\alpha}^g_{p,j}},\boldsymbol{\omega}'_{\boldsymbol{\alpha}^g_{p,j}})\}_{j=1,2}$ of $\{C_{\boldsymbol{\alpha}^g_{p,j}}\}_{j=1,2}$  on validation set $\mathcal{D}_{vld}$.
Finally, $\Omega$ is updated by $\{\boldsymbol{\omega}'_{\boldsymbol{\alpha}^g_{p,j}}\}_{j=1,2}$ as:
\begin{equation}\label{eq:update_weight_set}
	\Omega = \boldsymbol{\omega}'_{\boldsymbol{\alpha}^g_{p,f}} \cup\{\boldsymbol{\omega}'_{\boldsymbol{\alpha}^g_{p,s}} - \boldsymbol{\omega}'_{\boldsymbol{\alpha}^g_{p,f}} \cap \boldsymbol{\omega}'_{\boldsymbol{\alpha}^g_{p,s}}\} \cup \{\Omega-\boldsymbol{\omega}'_{\boldsymbol{\alpha}^g_{p,f}} \cup \boldsymbol{\omega}'_{\boldsymbol{\alpha}^g_{p,s}}\},
\end{equation}
where $\boldsymbol{\omega}'_{\boldsymbol{\alpha}^g_{p,f}}$ and $\boldsymbol{\omega}'_{\boldsymbol{\alpha}^g_{p,s}}$ are weights from fast-learner and slow-learner (refer to Section~\ref{arch_update}), respectively.
The first term $\boldsymbol{\omega}'_{\boldsymbol{\alpha}^g_{p,f}}$ means that $\Omega$ receives all the weights from $\omega'^g_{p,f}$ as it is assumed that $\omega'^g_{p,f}$, as the weights of fast-learner, is generally more valuable than $\omega'^g_{p,s}$.
The second term $\{\boldsymbol{\omega}'_{\boldsymbol{\alpha}^g_{p,s}} - \boldsymbol{\omega}'_{\boldsymbol{\alpha}^g_{p,f}} \cap \boldsymbol{\omega}'_{\boldsymbol{\alpha}^g_{p,s}}\}$ means that $\Omega$ receives the weights in $\boldsymbol{\omega}'_{\boldsymbol{\alpha}^g_{p,s}}$ but not in $\boldsymbol{\omega}'_{\boldsymbol{\alpha}^g_{p,f}}$.
The third term $\{\Omega-\boldsymbol{\omega}'_{\boldsymbol{\alpha}^g_{p,f}} \cup \boldsymbol{\omega}'_{\boldsymbol{\alpha}^g_{p,s}}\}$ mean that $\Omega$ keeps those unused weights unchanged.

With the above procedure as further illustrated in Fig.~\ref{fig:arch_eval}, it is expected that the weight set $\Omega$ becomes increasingly knowledgeable by co-evolving with the candidate architectures during the iterative slow-fast learning process, such that the performance estimation strategy is able to save substantial computation costs.

\section{Experiments}\label{sec:result}
In this section, we first provide the basic experiment on architecture search using RelativeNAS.
Then, we elaborate two analytical experiments to investigate some core properties of RelativeNAS.
Afterwards, we present experimental results on CIFAR-$10$ to evaluate the discovered cell architectures and comparisons with other state-of-the-art networks.
Finally, we show the transferability of our discovered cell architectures in both intra- and inter-tasks. 

\begin{figure*}[!htb]
     \centering
     \includegraphics[width=1\linewidth]{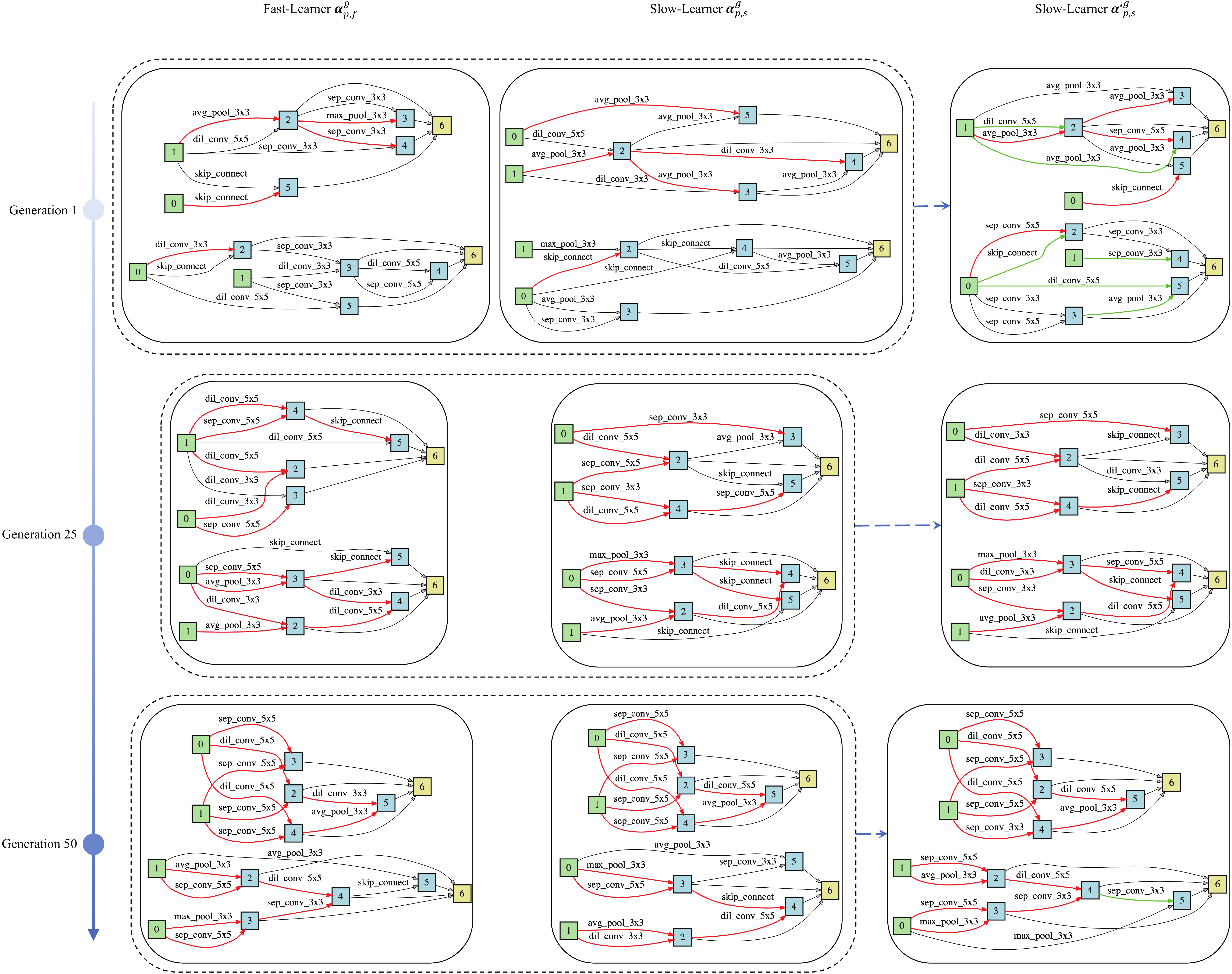}
     \caption{Illustration of the slow-fast learning with decoded architectures at generation $1$, $25$, and $50$ respectively. 
     Slow-learner updates its connections and operations by learning from fast-learner.
      The {\color{red} red lines} denote the common connections between fast-learner and slow-learner, and the {\color{green} green lines} denote the new connections after learning.}\label{fig:learning_data}

\end{figure*}

\subsection{Architecture Search}\label{ssec:exp_setting}
Basically, this work performs NAS on CIFAR-$10$~\cite{krizhevsky2009learning} which is widely used for benchmarking image classification.
Specifically, CIFAR-$10$ contains 60K images with a spatial solution of $32 \times 32$ and these images are equally divided into 10 classes, where the training set and the testing set are 50K and 10K, respectively.
Half of the training images of CIFAR-$10$ are randomly taken out as the search validation set. 

In RelativeNAS, the population size $N$ and the generation number are set to $20$ and $50$, respectively.
To evaluate the discovered architectures, every architecture vector $\boldsymbol{\alpha}$ is first decoded into a small network with $8$ cells (\emph{i.e.} $s = 2$) and initial channels set to 16. 
Then, this work trains those networks on the training set for one epoch using SGD with the weight decay and batch size set to $3 \times 10^{-4}$ and $256$ respectively.
In addition, the initial learning rate lr is $0.1$ which decays to zero following a cosine annealing schedule with $T_{max}$ set to the generation number, and Cutout~\cite{devries2017improved} of length 16 and the path dropout~\cite{ghiasi2018dropblock} with the probability of 0.3 are both applied for regularization.
The trained networks are assessed on the validation set to distinguish fast-learners and slow-learner by comparing the validation losses.
All in all, it takes about~\textbf{nine hours} with~\textbf{a unique 1080Ti} or~\textbf{seven hours} with~\textbf{a Tesla V100} to complete the above search procedure. 

\newcommand{\tabincell}[2]{\begin{tabular}{@{}#1@{}}#2\end{tabular}} 

\begin{table*}[!htbp]
\centering
\caption{Comparisons with other state-of-the-art methods on CIFAR-$10$.}
\begin{tabular}{lcccccc}
\hline
Architecture & \tabincell{c}{ Test Error \\ (\%)} & \tabincell{c}{Params \\ (M)} & \tabincell{c}{Search Cost \\ (GPU days)} & \tabincell{c}{GPU \\ Device} & \tabincell{c}{Search \\ Method} \\
\hline
FractalNet~\cite{larsson2016fractalnet} & 5.22 & 38.6 & - & - & manual \\
Wide-ResNet~\cite{zagoruyko2016wide} & 4.17 & 36.5 & - & - & manual \\
DenseNet-BC~\cite{huang2017densely} & 3.46 & 25.6 & - & - & manual \\
\hline
PNAS~\cite{liu2018progressive} & 3.41 & 3.2 & 225 & - & SMBO \\
NAONet + Cutout~\cite{luo2018neural} & 2.48 & 10.6 & 200 & V100 & NAO \\
Random search + Cutout~\cite{liu2018darts} & 3.29  & 3.2 & 4 & 1080Ti & random \\
DARTS(first) + Cutout~\cite{liu2018darts} & 3.0  & 3.3 & 1.5 & 1080Ti & gradient-based \\

DARTS(second) + Cutout~\cite{liu2018darts} & 2.76  & 3.3 & 4.0 & 1080Ti & gradient-based \\

\tabincell{c}{SNAS+mild constraint + Cutout~\cite{xie2018snas}} & 2.98  & 2.9 & 1.5 & TITAN Xp & gradient-based \\

\tabincell{c}{SNAS+moderate constraint + Cutout~\cite{xie2018snas}} & 2.85  & 2.8 & 1.5 & TITAN Xp & gradient-based \\

\tabincell{c}{SNAS+aggressive constraint + Cutout~\cite{xie2018snas}} & 3.10 & 2.3 & 1.5 & TITAN Xp & gradient-based \\

ProxylessNAS + Cutout~\cite{cai2018proxylessnas} & 2.08  & 5.7 & 4.0 & V100 & gradient-based 
\\
MetaQNN~\cite{baker2016designing} & 6.92 & 11.8 & 100 & - & RL \\  
NASNet-A + Cutout~\cite{zoph2018learning} & 2.65 & 3.41 & 2000 & K40 & RL \\
BlockQNN + Cutout~\cite{zhong2020blockqnn}  & 2.80 & 39.8 & 32 & 1080Ti & RL \\
ENAS + Cutout~\cite{pham2018efficient} & 2.89 & 4.6 & 0.5 & 1080Ti & RL \\
\hline
AmoebaNet-A~\cite{real2018regularized}  & 3.34 & 3.2 & 3150 & K40 & population-based \\
AmoebaNet-B + Cutout~\cite{real2018regularized} & 2.55 & 2.8 & 3150 & K40 & population-based \\
Large-Scale Evolution~\cite{real2017large} & 5.4 & 5.4 & 2600 & - & population-based \\
Hierarchical Evolution~\cite{liu2017hierarchical} & 3.75 & 15.7 & 300 & P100 & population-based \\
EffPnet~\cite{wang2021surrogate} & 3.49 & 2.54 & $<$3 & RTX 2070 & population-based \\

\hline
\textbf{RelativeNAS} + Cutout & \textbf{2.34} & \textbf{3.93} & \textbf{0.4} & \textbf{1080Ti} & \textbf{population-based} \\

\bottomrule
\end{tabular}
\label{tab:cifar10}
\end{table*}

\subsection{Population Searching Analysis}\label{ssec:pop_conv}

Fig.~\ref{fig:pop_data} (top) provides the validation losses of all the $20$ decoded candidate architectures in a population over the searching process.
Initially, the losses differ widely among the candidate architectures due to the randomly initialized architectures; as the search proceeds, the differences of the losses gradually decrease towards a stable value, indicating convergence of the population.

For more insightful observations, this work randomly selects one decoded candidate architecture in the population to trace its architectures obtained over the searching process.
As shown in Fig.~\ref{fig:pop_data} (bottom),
at the initial stage (generation $1$) of the searching process, the normal and reduction cells are randomly generated, without any specified property;
at the middle stage (generation $30$), however, the normal cell becomes denser while the reduction cell remains flat;
at the final stage (generation $50$), the topologies of the two types of cells remain stable, despite the changes in the detailed connections inside them.
%
Such observations indicate that the population searching process would generate expected candidate architectures towards converged optima, in terms of both connections and operations.
Indeed, this work further trains the architecture shown at the final stage on the different datasets to validate its performance.

\subsection{Slow-Fast Learning Analysis}

To empirically analyze the slow-fast learning process, this work randomly picks up three pairs of architectures obtained at generation $1$, generation $25$, and generation $50$, respectively, to provide the illustrative example in Fig.~\ref{fig:learning_data}.  
At generation $1$, the architectures of fast-learner and slow-learner both are randomly initialized at the beginning.
Hence, there exist substantial differences between fast-learner and slow-learner, such that slow-learner substantially changes its connections as well as operations after learning from fast-learner.
At generation $25$, there are some common connection patterns between fast-learner and slow-learner, \emph{e.g.} the two predecessor nodes of Node $4$ are both Node $1$ in the normal cells, and the two predecessor nodes of Node $3$ are both Node $0$ in the reduction cells.
Therefore, slow-learner will not change these patterns during slow-fast learning.
By contrast, due to the differences between operations Dil Conv $3\times3$ and Sep Conv $5\times5$ in the connection between Node $2$ and Node $0$ for fast-learner and slow-learner, slow-learner learns from fast-learner and changes to Dil Conv $3\times3$. 
At generation $50$, the connections of the normal cells are exactly the same, and thus slow-learner only makes some minor adjustments in its operations by learning from fast-learner. 
Besides, despite that there is still a gap between fast-learner and slow-learner in the reduction cells, while the overall architectures become quite similar after $50$ generations of slow-fast learning. 
Based on the above observations, this work can conclude that the slow-fast learning paradigm is generally effective over the search process, showing different functionalities at different stages.  



\begin{table*}[!htbp]
\centering
\caption{Comparison with other state-of-the-art networks on CIFAR-$100$. $\dagger$ denotes directly searching on CIFAR-$100$, while others are searched on CIFAR-$10$.}
\begin{tabular}{lccccc}
\hline
Architecture & \tabincell{c}{ Test Error \\ (\%)} & \tabincell{c}{Params \\ (M)} & \tabincell{c}{Search Cost \\ (GPU days)}  &  \tabincell{c}{Search \\ Method} \\
\hline
FractalNet~\cite{larsson2016fractalnet} & 23.30 & 38.6 & -  & manual \\
Wide-ResNet~\cite{zagoruyko2016wide} & 20.50 & 36.5 & - & manual \\
DenseNet-BC~\cite{huang2017densely} & 17.18 & 25.6 & -  & manual \\
\hline
NAONet + Cutout~\cite{luo2018neural} & 15.67 & 10.8 & 200  & NAO \\ 
MetaQNN~\cite{baker2016designing} & 27.14 & 11.8 & 100  & RL \\  
\hline
Large-Scale Evolution$^{\dagger}$~\cite{real2017large} & 23.0 & 40.4 & 2600  & population-based \\
EffPnet~\cite{wang2021surrogate} & 18.49 & 2.54 & $<$3  & population-based \\
\hline
\textbf{RelativeNAS} + Cutout  & \textbf{15.86} & \textbf{3.98} & \textbf{0.4}  & \textbf{population-based} \\

\bottomrule
\end{tabular}
\label{tab:cifar100}
\end{table*}

\begin{table*}[!htbp]
\centering
\caption{Comparison with other state-of-the-art methods on ImageNet. $\dagger$ denotes directly searching over ImageNet, while others are searched on CIFAR-$10$.}
\begin{tabular}{lcccccc}
\hline
 \multirow{2}{*}{Architecture} &  \multicolumn{2}{c}{Test Error (\%)} & Params  & $\times$ $+$  &Search Cost &  Search \\
 \cline{2-3}
& top-$1$  & top-$5$ &  (M) &  (M) & (GPU days) & Method\\
\hline
Inception-V1~\cite{szegedy2015going} & 30.2 & 10.1 & 6.6 & 1448 & -  & manual \\
MobileNet-V1 (1x)~\cite{howard2017mobilenets} & 29.4 & 10.5 & 4.2 & 575 & - &  manual \\
MobileNet-V2 (1.4)~\cite{sandler2018mobilenetv2} & 25.3 & - & 6.9 & 585 & - &  manual \\
ShuffleNet-V1 (2x)~\cite{zhang2018shufflenet} & 26.4 & 10.2 & ~5 & 524 & - &  manual \\
ShuffleNet-V2 (2x)~\cite{ma2018shufflenet} & 25.1 & - & ~5 & 591 & - & manual \\

\hline
PNAS~\cite{liu2018progressive} & 25.8 & 8.1   & 5.1   & 588   & 225   &  SMBO \\
NAONet~\cite{luo2018neural} & 25.7 & 8.2 & 11.35 & 584 & 200 &  NAO \\ 
DARTS (second order)~\cite{liu2018darts} & 26.7 & 8.7 & 4.7 & 574 & 4.0 &  gradient-based \\
SNAS (mild constraint)~\cite{xie2018snas} & 27.3 & 9.2 & 4.3 & 533 & 1.5 &  gradient-based \\
ProxylessNAS (GPU)$^{\dagger}$~\cite{cai2018proxylessnas}  & 24.9 & 7.5 & 7.1 & 465 & 8.3 &  gradient-based \\

NASNet-A~\cite{zoph2018learning} & 26.0 & 8.4 & 5.3 & 564 & 1800 & RL \\
NASNet-B~\cite{zoph2018learning} & 27.2 & 8.7 & 5.3 & 488 & 1800 & RL \\
NASNet-C~\cite{zoph2018learning} & 27.5 & 9.0 & 4.9 & 558 & 1800 & RL \\
\hline

AmoebaNet-A~\cite{real2018regularized} & 25.5 & 8.0 & 5.1 & 555 & 3150 & population-based \\
AmoebaNet-B~\cite{real2018regularized} & 26.0 & 8.5 & 5.3 & 555 & 3150 & population-based \\
AmoebaNet-C~\cite{real2018regularized} & 24.3 & 7.6 & 6.4 & 570 & 3150 & population-based \\ 
EffPnet~\cite{wang2021surrogate} & 27.01 & 9.25 & 2.54 & - & $<$3 & population-based \\

\hline

\textbf{RelativeNAS}   & \textbf{24.88} & \textbf{7.7} & \textbf{5.05} & \textbf{563} &\textbf{0.4}& \textbf{population-based} \\

\bottomrule
\end{tabular}
\label{tab:imagenet}
\end{table*}

\subsection{Results on CIFAR-10}\label{ssec:res_cifar10}

In this work, we follow the training settings of DARTS~\cite{liu2018darts} for image classification tasks to train networks from randomly initialized weights, for fair comparisons between the searched architectures.
A large network of $20$ cells (\emph{i.e.} $s$ is set to $6$) is built with the selected normal and reduction cells while the initial number of channels is set to $36$. 
Most hyper-parameters are the same as the ones used in the above search process except lr, path dropout, and batch size which are set to $0.025$, $0.2$, and $128$, respectively. 
For further enhancement, an auxiliary head with weight $0.4$ is added into the network.
Instead of half training images, this work trains the network from scratch over the whole training set for 600 epochs and evaluate it over the test set. 

The results and comparisons with other state-of-the-art networks (including manual and NAS) on CIFAR-$10$ are summarized in Table~\ref{tab:cifar10}.
Besides, we label all the GPU devices for different NAS methods.
Compared with the manual networks, our RelativeNAS has fewer parameters while outperforms them by a large margin.
The proposed RelativeNAS gains an encouraging improvement to DARTS~\cite{liu2018darts} and a random search baseline~\cite{liu2018darts} in terms of test error and search cost. 
Compared with other NAS networks, it can be observed that ours needs the least cost on time while gets superior results in terms of test error and parameters. 
Although ProxylessNAS~\cite{cai2018proxylessnas} achieves less test error than ours ($2.08\%$ vs $2.34\%$), it has much more parameters ($5.7$M vs $3.9$M) and costs $10\times$ longer search time than ours ($4$ vs $0.4$).
While EffPnet~\cite{wang2021surrogate} has fewer parameters ($2.54$M vs $3.9$M), the test error and search cost are larger than ours. 
Furthermore, RelativeNAS is the only one involving the pseudo gradient between architecture vectors among those population-based NAS.
To the best of our knowledge, our RelativeNAS is the most efficient search method among those population-based methods.
With ENAS~\cite{pham2018efficient} and RelativeNAS proposed, RL-based and population-based methods are no longer time-consuming and even $2\times$ faster than gradient-based methods.
Moreover, the proposed RelativeNAS outperforms ENAS in all aspects.

\subsection{Transferability Analyses}\label{ssec:transfer}
In this subsection, we will validate the transferability of the normal and reduction cells discovered by the proposed RelativeNAS on CIFAR-$10$. 
We first validate their generality in other image classification tasks (\emph{i.e.} intra-tasks), and then demonstrate their transferability in inter-tasks, including object detection, semantic segmentation, and keypoint detection.  

\subsubsection{Intra-task Transferability}
The discovered normal cell and reduction cell on CIFAR-$10$ both are directly transferred to CIFAR-$100$ and ImageNet without further search. 
Since the architecture is transferred, the overall search cost is the same as on CIRAR-$10$.

\textbf{CIFAR-100.} CIFAR-$100$ contains 60K images with a spatial resolution of $32 \times 32$, where 50K images are used as the training set and the left 10K images are used as the testing set.
Moreover, these images are distributed equally for 100 classes.
The network used in CIFAR-$10$ is directly transferred to CIFAR-$100$ with a small modification in the last classification layer to adapt to the different number of classes.
The training details are the same as CIFAR-$10$ except the weight decay and batch size which is set to $5\times10^{-4}$ and $96$, respectively.

Table~\ref{tab:cifar100} shows the experimental results and comparisons with other state-of-the-art networks.
Surprisingly, our direct transferred network achieves a test rate of $15.86\%$ and still outperforms most networks.
In particular, RelativeNAS outperforms Large-Scale Evolution~\cite{real2017large} by about $7$ points which is searching on CIFAR-$100$ instead of transferring from CIFAR-$10$.   
It can be concluded that our RelativeNAS derived from CIFAR-$10$ is indeed transferable to a more complicated task (\emph{i.e.} CIFAR-$100$) while maintains its superiority. 

\textbf{ImageNet.}
ImageNet 2012 dataset~\cite{deng2009imagenet} is one of the most challenging benchmarks for image classification, which is consisted of 1.28M and 50K images for training and validation, respectively. 
Those images are unevenly distributed in the $1000$ different classes, and they do not have unified spatial resolution but usually much larger than $32\times32$. 
In order to fit such a difficult dataset, this work follows the common practice~\cite{liu2018darts} to modify the network structure used in CIFAR-$10/100$. 
To be more concrete, the macro-architecture starts with three convolutional layers with stride set to 2, which can reduce the spatial resolution of input images $8$ times. 
In the following, $14$ cells (\emph{i.e.} $s$ is set to $4$) are stacked. 
With the consideration of the mobile setting (\emph{i.e.} the number of multiply-add operations should be less than 600M), the initial channel is set to 46. 
This work trains the model over the train set while reporting the results on the validation set. 
During training, this work adopts some common data augmentation strategies, including randomly resize and crop, random horizon flip, and color jitter.
There are $1024$ examples in each training batch and the size of each image is equal to $224\times224$.
The model is optimized with SGD for $250$ epochs, where the initial learning rate, momentum, and weight decay are set to $0.5$, $0.9$, and $3\times10^{-5}$, respectively.
This work applies the warm-up strategy over the first $5$ epochs, where the learning rate is gradually increasing linearly from $0$ to the initial value. 
During the left $245$ epochs, the learning rate decays linearly from $0.5$ to $1\times10^{-5}$.
In addition, this work also uses the label smoothing~\cite{szegedy2016rethinking} to regularize our model, and the smoothing parameter $\epsilon$ is equal to 0.1.

The results of RelativeNAS compared with other state-of-the-art networks on the ImageNet are presented in Table~\ref{tab:imagenet}.
It is worth noticing that RelativeNAS achieves competitive performance, \emph{i.e.}top-$1$ and top-$5$ test error rate of $24.88\%$ and $7.7\%$, respectively.
Interestingly, our transferred RelativeNAS performs a little better than ProxylessNAS (GPU)~\cite{cai2018proxylessnas} which is searching on ImageNet directly. 
The results further demonstrate that the proposed method enables the transformation of simple cells to complex macro architectures for solving more complicated tasks with low cost but high performance.

\subsubsection{Inter-task Transferability}
We will further demonstrate the transferability by transferring our network pretrained on ImageNet to other tasks instead of image classification. 
To be more specific, we will train and evaluate SSD~\cite{liu2016ssd}, BiSeNet~\cite{yu2018bisenet}, and SimpleBaseline~\cite{xiao2018simple} with different mobile-setting backbones under the same training settings for object detection, semantic segmentation, and keypoint detection, respectively. 
We note that all compared models (network structures as well as the ImageNet pretrained weights) in this part are from \href{https://pytorch.org/docs/stable/torchvision/models.html}{PyTorch repository} except DARTS~\cite{liu2018darts}, which is from \href{https://github.com/quark0/darts}{the official released GitHub repository}. 

\begin{figure*}[!htb]
     \centering
     \includegraphics[width=0.98\linewidth]{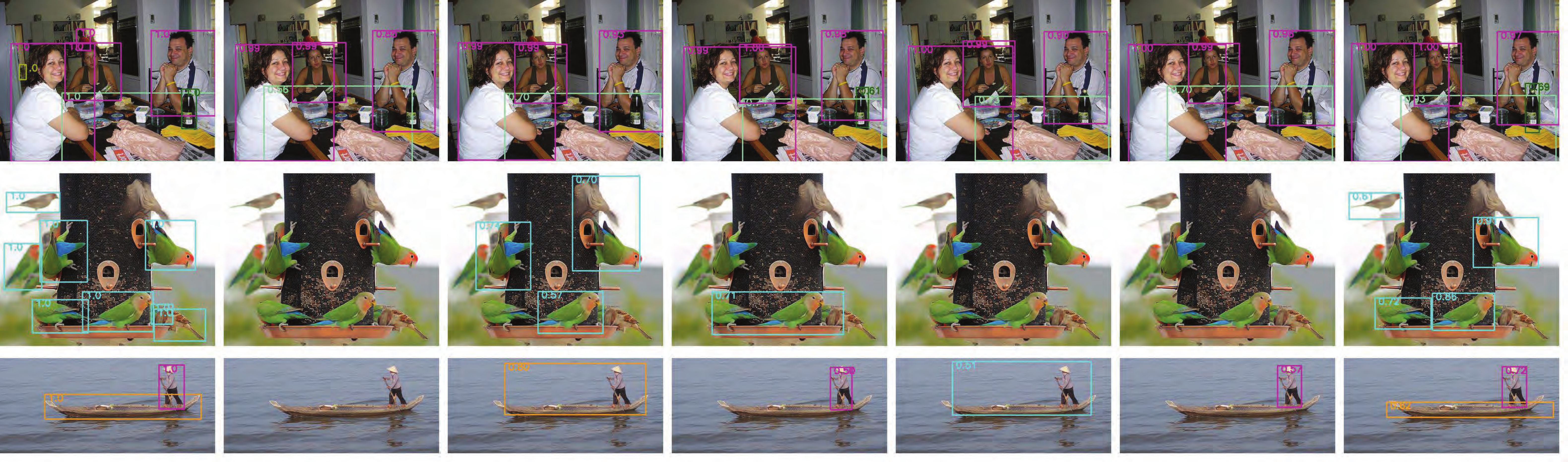} 
     \caption{Visual examples achieved by SSDLite with different backbones.
        From left to right are groudtruth, ShuffleNetV2, MobileNetV2, NASNet, MnasNet, DARTS, and RelativeNAS, respectively. The confidence threshold is 0.5. Different colors represent different classes.} \label{fig:ssdlite_visual_result}
\end{figure*}

\textbf{Object Detection.}
For object detection, this work compares our network with other counterparts on PASCAL VOC, in which thousands of images over 20 object classes are annotated with bounding boxes.
Among those object classes, bottles and plants are both small objects. 
Following~\cite{sandler2018mobilenetv2}, this work adopts the SSDLite as our object detection framework, which is a mobile-friendly variant of Single Shot Detector (SSD)~\cite{liu2016ssd}.
Specifically, all the regular convolutions are replaced with separable convolutions in SSD extra layers and prediction layers, with which SSDLite is slighter and more efficient than the original SSD. 
This work trains all models over the combined trainval sets of VOC 2007 and 2012 using SGD with a batch size of 32, the momentum of $0.9$, and weight decay of $5\times10^{-4}$.
Besides, input images are resized to $320 \times 320$ and the learning rate is set to $0.01$ which will decay to zero in 200 epochs with cosine annealing scheduler without a restart.
Table~\ref{tb:obj_det_voc} presents the performance achieved by those models on PASCAL VOC 2007 test set. 
This work can conclude that our RelativeNAS achieves the best performance while keeps comparable in terms of parameters and FLOPs under the same settings.
Moreover, the discovered model outperforms others in small objects by a large margin, which can be attributed to the fact that our model has a strong ability to retain spatial details while extracting abstract semantic information. 
In addition to the quantitative comparison, this work also provides some qualitative results in Fig.~\ref{fig:ssdlite_visual_result}.
From it, this work can see that our model indeed surpasses others in detecting the bottle (first row) and bird (second row). 
Furthermore, it seems our model can well exploit the surrounding context to improve performance, as it can identify this is a boat instead of a bird in the last row. 

\begin{table*}[!htbp]
\centering
\caption{Results of SSDLite~\cite{sandler2018mobilenetv2} with different mobile-setting backbones on PASCAL VOC 2007 test set.}
\label{tb:obj_det_voc}
\begin{threeparttable}
\begin{tabular}{l|c|c|c|c|c}
\hline
\multirow{2}{*}{Backbone}& \multirow{2}{*}{\# Params (M)} & \multirow{2}{*}{\# FLOPs (M)} & \multicolumn{2}{c|}{Small Objets (AP (\%))}  & \multirow{2}{*}{mAP (\%)} \\ \cline{4-5}
                                                    &               &               & Bottle    & Plant     &   \\
\hline
ShuffleNet-V2 (1x)~\cite{ma2018shufflenet}          & 2.17          & 355.76        & 29.9      & 38.1      & 65.4   \\
MobileNet-V2 (1x)~\cite{sandler2018mobilenetv2}     & 3.30          & 680.88        & 37.9      & 43.9      & 69.4   \\
\hline 
NASNet~\cite{zoph2018learning}                      & 5.22          & 1238.92       & 41.5      & 46.1      & 71.6   \\
MnasNet~\cite{tan2019MnasNet}                       & 4.18          & 708.72        & 37.7      & 44.4      & 69.6  \\
DARTS~\cite{liu2018darts}                           & 4.73          & 1138.16       & 38.3      & 49.3      & 71.2   \\
\textbf{RelativeNAS}           & \textbf{5.07}  & \textbf{1202.97}  & \textbf{45.9}   & \textbf{50.3}   & \textbf{73.1}   \\
\hline
\end{tabular}
\end{threeparttable}
\end{table*}

\begin{figure*}[!htb]
     \centering
     \includegraphics[width=0.98\linewidth]{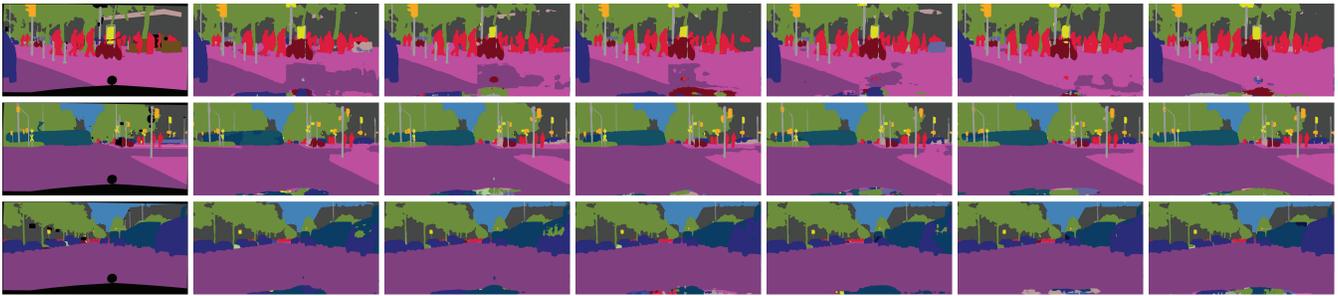} 
     \caption{Visual examples achieved by BiSeNet with different backbones.
        From left to right are ground truth, ShuffleNetV2, MobileNetV2, NASNet, MnasNet, DARTS, and RelativeNAS, respectively. Different colors denote different classes.} \label{fig:semseg_visual_result}
\end{figure*}

\textbf{Semantic Segmentation.}
Cityscapes~\cite{cordts2016cityscapes} is a large-scale dataset containing pixel-level annotations of 5000 images (2975, 500, and 1525 for the training, validation, and test sets respectively) and about 20000 coarsely annotated images. 
Following the evaluation protocol~\cite{cordts2016cityscapes}, $19$ semantic labels are used for evaluation without considering the void label.
This work evaluates the BiSeNet~\cite{yu2018bisenet} with different mobile-setting backbones on the validation set when training with only 2975 images (\emph{i.e.}train fine set). 
All models are trained for 80K iterations with the initial learning rate and batch size set to $1\times10^{-2}$ and $16$, respectively.
Similar to~\cite{yu2018bisenet}, this work decays the lr with the ''poly'' learning rate strategy.
To be more concrete, the initial lr is multiplied by $(1 - \frac{iter}{max\_iter})^{0.9}$.
This work follows the BiSeNet to augment our training images. Specifically, this work employs the color jitter, random scale ($scales = \{0.75, 1, 1.25, 1.5, 1.75, 2.0\}$), and random horizontal flip. After that, this work randomly crops the augmented images into a fixed size that is $1024 \times 1024$ for training.
Note that, the multi-crop testing is adopted during the test phase, and the test crop size is equal to $1024 \times 1024$, too.
Table ~\ref{tb:sem_seg_city} provides the comparison with several representative mobile-setting backbones on the Cityscapes val set in terms of the parameter, computation complexity (\emph{i.e.} FLOPs) and mIoU. 
It can be seen that our RelativeNAS has fewer parameters and FLOPs than the NASNet while outperforming NASNet by 0.8 points in terms of mIoU.
Furthermore, when compared with BiSeNet that adopts ResNet101 as the backbone~\cite{yu2018bisenet}, our RelativeNAS achieves a better result over val set ($80.4$ $vs$ $80.3$) when adopting the same multi scales ($scales = \{0.5, 0.75, 1, 1.25, 1.5, 1.75\}$) as well as flipping during inference. 
Some visual examples are displayed in Fig.~\ref{fig:semseg_visual_result}, where it can be seen that the BiSeNet paired with our RelativeNAS can better segment the boundaries of objects.   
\begin{table}[!htbp]
\centering
\caption{Results of BiSeNet~\cite{yu2018bisenet} with different mobile-setting backbones on Cityscapes val set. (single scale and no flipping).}
\label{tb:sem_seg_city}
\begin{threeparttable}
\begin{tabular}{l|c|c|c}
\hline
Backbone   & \# Params (M) & \# FLOPs (G) & mIoU (\%) \\
\hline
ShuffleNet-V2 (1x)~\cite{ma2018shufflenet}                    & 4.10                  & 26.30         & 73.0      \\
MobileNet-V2 (1x)~\cite{sandler2018mobilenetv2}               & 5.24                  & 29.21         & 77.1      \\
\hline  
NASNet~\cite{zoph2018learning}                          & 7.46                  & 36.51         & 77.9      \\
MnasNet~\cite{tan2019MnasNet}                           & 6.12                  & 29.50         & 76.8      \\     
DARTS~\cite{liu2018darts}                               & 6.64                  & 34.77         & 77.5      \\
\textbf{RelativeNAS}                                & \textbf{6.94}        & \textbf{35.35}         & \textbf{78.7}      \\
\hline
\end{tabular}
\end{threeparttable}
\end{table}

\begin{figure*}[!htb]
     \centering
     \includegraphics[width=0.98\linewidth]{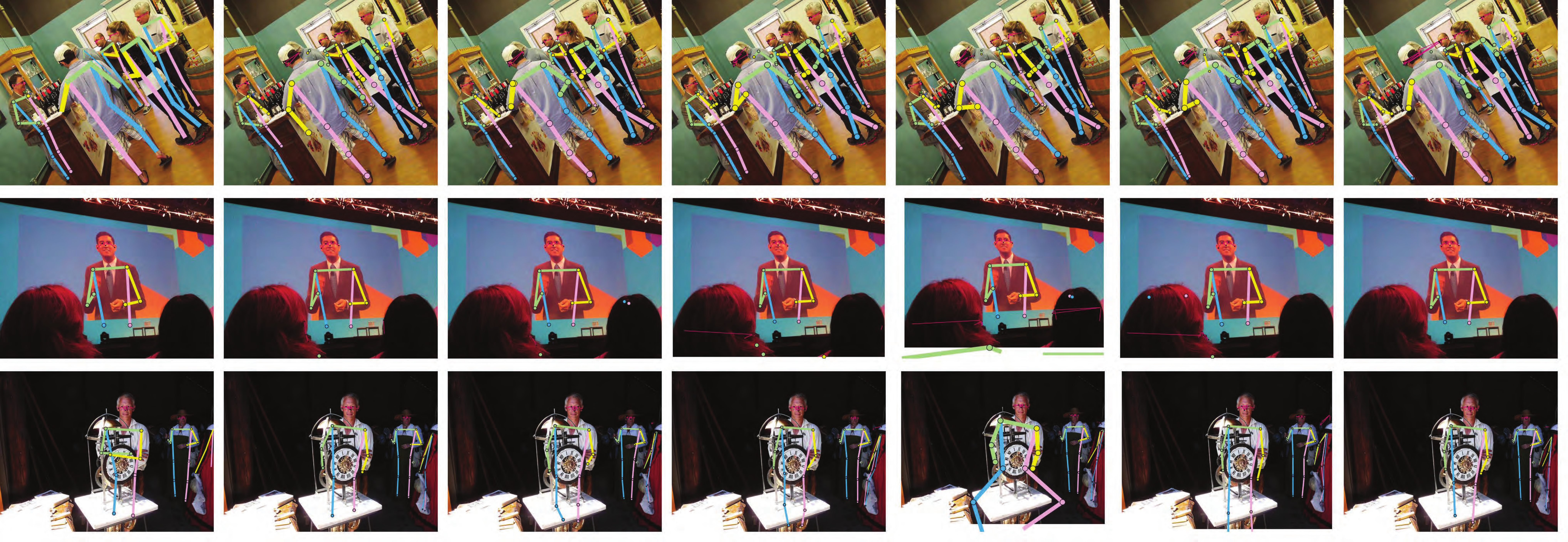} 
     \caption{Visual examples achieved by SimpleBaseline with different backbones.
        From left to right are ground truth, ShuffleNetV2, MobileNetV2, NASNet, MnasNet, DARTS, and RelativeNAS, respectively. 
        Different colors represent different keypoints.} \label{fig:simplebaseline_visual_result}
\end{figure*}

\textbf{Keypoint Detection.} Keypoint detection aims to detect the locations of $k$ human parts (\emph{e.g.}, ankle, shoulder, etc) from an image. 
The MSCOCO~\cite{lin2014microsoft} is a widely used benchmark dataset for keypoint detection which includes over 250k person instances labelled with 17 keypoints.  
SimpleBaseline~\cite{xiao2018simple} is adopted as our general keypoint detection framework, and this work assesses it when paired with different backbones. 
This work trains all models on the MSCOCO train2017 set and evaluate them on the val2017 set, containing 57K and 5K images, respectively. 
Following the SimpleBaseline~\cite{xiao2018simple}, this work crops the human detection boxes from the images which are then resized to $256 \times 192$.
In addition, the random rotation, scale, and flipping are all applied to data augmentation. 
Each model is trained with the Adam optimizer~\cite{kingma2014adam} for $140$ epochs and the initial learning rate is set to $1\times10^{-3}$ which will drop to $1\times10^{-4}$ and $1\times10^{-5}$ at the 90th and 120th epoch, respectively.
Moreover, each batch contains 128 examples. 
Similar as~\cite{xiao2018simple}, a two-stage top-down paradigm is adopted during inference.
To be more specific, an independent person detector is applied to detect the person instances and then those instances are input to the trained keypoint detector for predicting human keypoints.
This work reports the experimental results of SimpleBaseline with different mobile-setting backbones in Table~\ref{tb:key_det_coco}. 
Our RelativeNAS still performs better than others in terms of AP while is comparable in the other two aspects.
Moreover, our claims are supported by visual examples in Fig.~\ref{fig:simplebaseline_visual_result}.

\begin{table}[!htbp]
\centering
\caption{Results of SimpleBaseline~\cite{xiao2018simple} with different mobile-setting backbones on MS COCO2017 val set. Flip is used during validation.}
\label{tb:key_det_coco}
\begin{threeparttable}
\begin{tabular}{l|c|c|c}
\hline
Backbone    & \# Params (M)     & \# FLOPs (M)      & AP (\%)     \\
\hline
ShuffleNet-V2 (1x)~\cite{ma2018shufflenet}                & 7.55       & 154.37           & 60.4     \\
MobileNet-V2 (1x)~\cite{sandler2018mobilenetv2}           & 9.57       & 306.80           & 649     \\
\hline
NASNet~\cite{zoph2018learning}                      & 10.66      & 569.11           &   67.9   \\
MnasNet~\cite{tan2019MnasNet}                       & 10.45      & 320.17           & 62.5     \\
DARTS~\cite{liu2018darts}                           & 9.20       & 531.77           & 66.9    \\
\textbf{RelativeNAS}                        & \textbf{9.43}       & \textbf{564.19}           & \textbf{68.3}     \\

\hline
\end{tabular}
\end{threeparttable}
\end{table}

\section{Conclusions}\label{sec:conclusion}
 This paper has presented a framework, called RelativeNAS, for the effective and efficient automatic design of high-performance networks. 
Within RelativeNAS, a novel continuous encoding scheme for cell-based search space has been proposed firstly.
To further utilize the continuously encoded search space, a slow-fast learning paradigm has been applied as an optimizer to iteratively update the architecture vectors.
In contrast to existing learning/optimization methods in NAS, the proposed one does not directly use \emph{loss-based knowledge} to update the architectures.
Instead, the candidate architectures are made to learning from each other by the pariwisely generated pseudo-gradients, \emph{i.e.} slow-learner learning from fast-learner in each pair of candidate architectures. 
In addition, a performance estimation strategy has been proposed to reduce the cost of evaluating candidate architectures.
The effectiveness of such a strategy can be largely attributed to the fact that the validation loss is merely used for distinguishing slow-learner and fast-learner by partial ordering, which only requires estimated (instead of exact) loss values.

With the proposed RelativeNAS as above, consequently, it takes about nine 1080Ti GPU hours (\emph{i.e.} $0.4$ GPU Day) for our RelativeNAS to search on CIFAR-$10$. 
%
%
Furthermore, our discovered network has been able to outperform or match other state-of-the-art manual and NAS networks on CIFAR-$10$ while showing promising transferability in other intra- and inter-tasks, such as ImageNet, object detection. 
In particular, our transferred network has yielded the best performance on PASCAL VOC, Cityscapes, and MS COCO. 
In conclusion, this work highlights the merits of differentiable NAS and combining population-based NAS, to be more effective and more efficient.
Moreover, the proposed slow-fast learning paradigm can be also potentially applicable to other generic learning/optimization tasks.


%

\section*{Acknowledgment}
This work was supported by the National Natural Science Foundation of China (No. 61903178, 61906081, and U20A20306), the Shenzhen Science and Technology Program (No. RCBS20200714114817264), the Program for Guangdong Introducing Innovative and Entrepreneurial Teams (No. 2017ZT07X386), the Shenzhen Peacock Plan (No. KQTD2016112514355531), the Program for University Key Laboratory of Guangdong Province (No. 2017KSYS008), and the General Research Fund of Hong Kong (No. 27208720). 
The authors are grateful to Zhichao Lu for his comments and support on the paper.

\bibliography{ref}
\bibliographystyle{IEEEtran}

\begin{IEEEbiography}[{\includegraphics[width=1in,height=1.2in,clip,keepaspectratio]{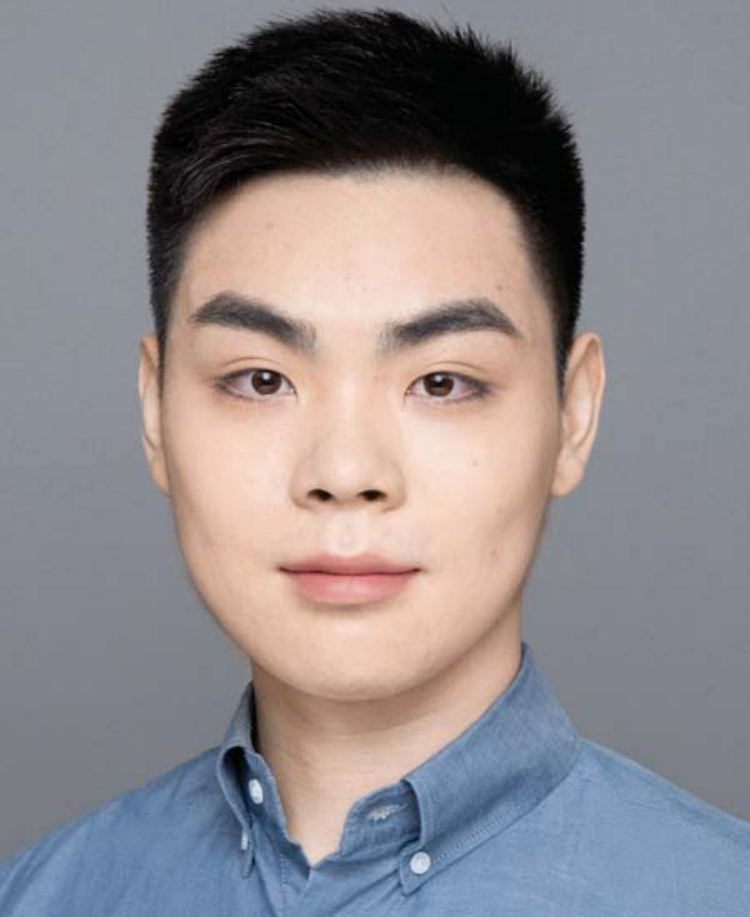}}]{Hao Tan} 
received the B.Eng. degree in computer science and technology from the Southern University of Science and Technology, Shenzhen, China, in 2020.
He is currently a Research Assistant with the Department of Computer Science and Engineering, Southern University of Science and Technology, Shenzhen, China. His current research interests include computer vision, neural architecture search, and swarm intelligence.
\end{IEEEbiography}
\begin{IEEEbiography}[{\includegraphics[width=1in,height=1.2in,clip,keepaspectratio]{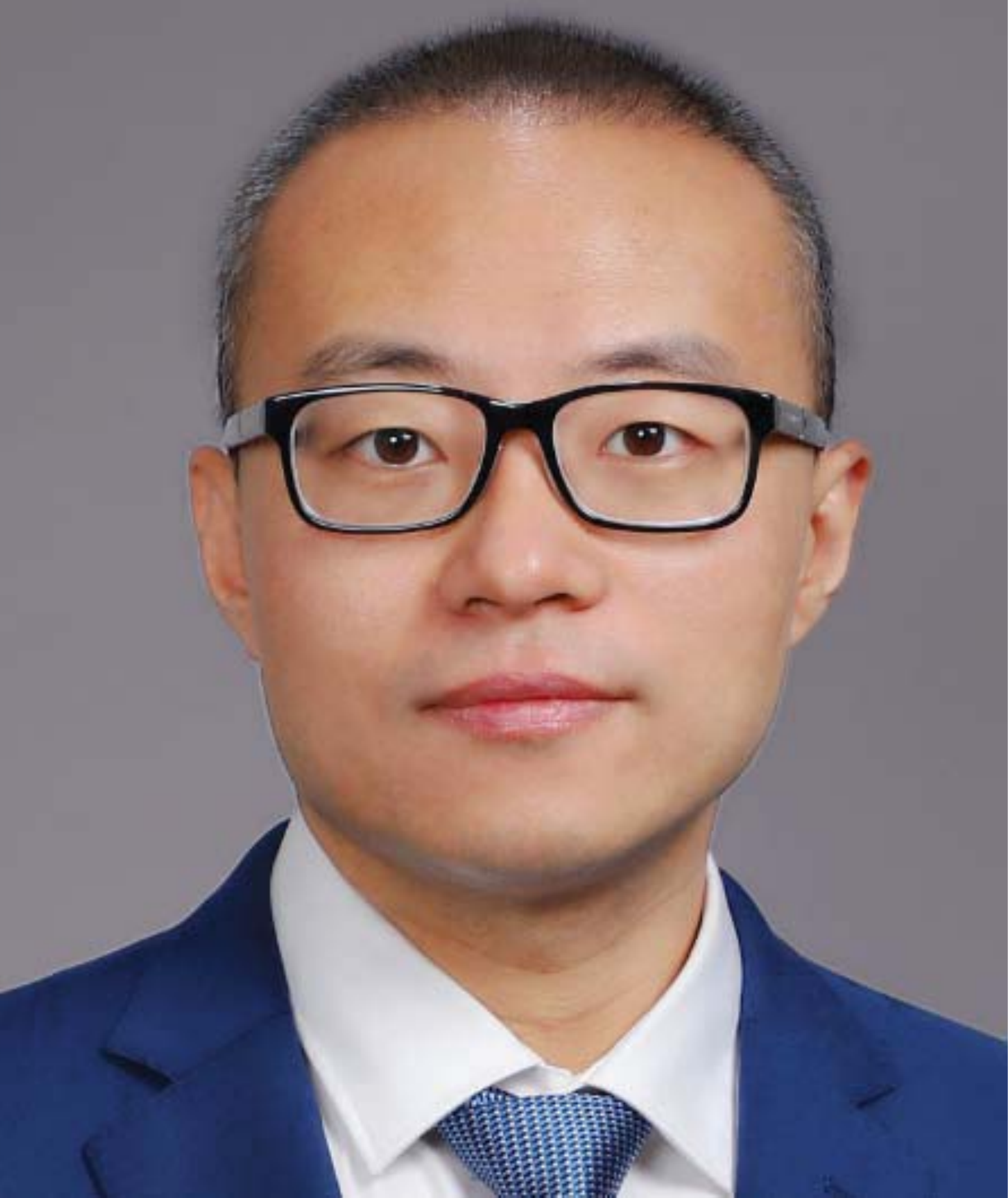}}]{Ran Cheng} (Senior Member, IEEE)
 received the B.Sc. degree in computer science and technology from the Northeastern University, Shenyang, China, in 2010, and the Ph.D. degree in computer science from the University of Surrey, Guildford, U.K., in 2016.
He is currently an Associate Professor with the Department of Computer Science and Engineering, Southern University of Science and Technology, Shenzhen, China. Dr. Cheng was a recipient of the IEEE TRANSACTIONS ON EVOLUTIONARY COMPUTATION Outstanding Paper Award in 2018 and 2021, the IEEE Computational Intelligence Society Outstanding Ph.D. Dissertation Award in 2019, and the IEEE COMPUTATIONAL INTELLIGENCE MAGAZINE Outstanding Paper Award in 2020. He is an Associate Editor of the IEEE TRANSACTIONS ON ARTIFICIAL INTELLIGENCE.
 
\end{IEEEbiography}
\begin{IEEEbiography}[{\includegraphics[width=1in,height=1.2in,clip,keepaspectratio]{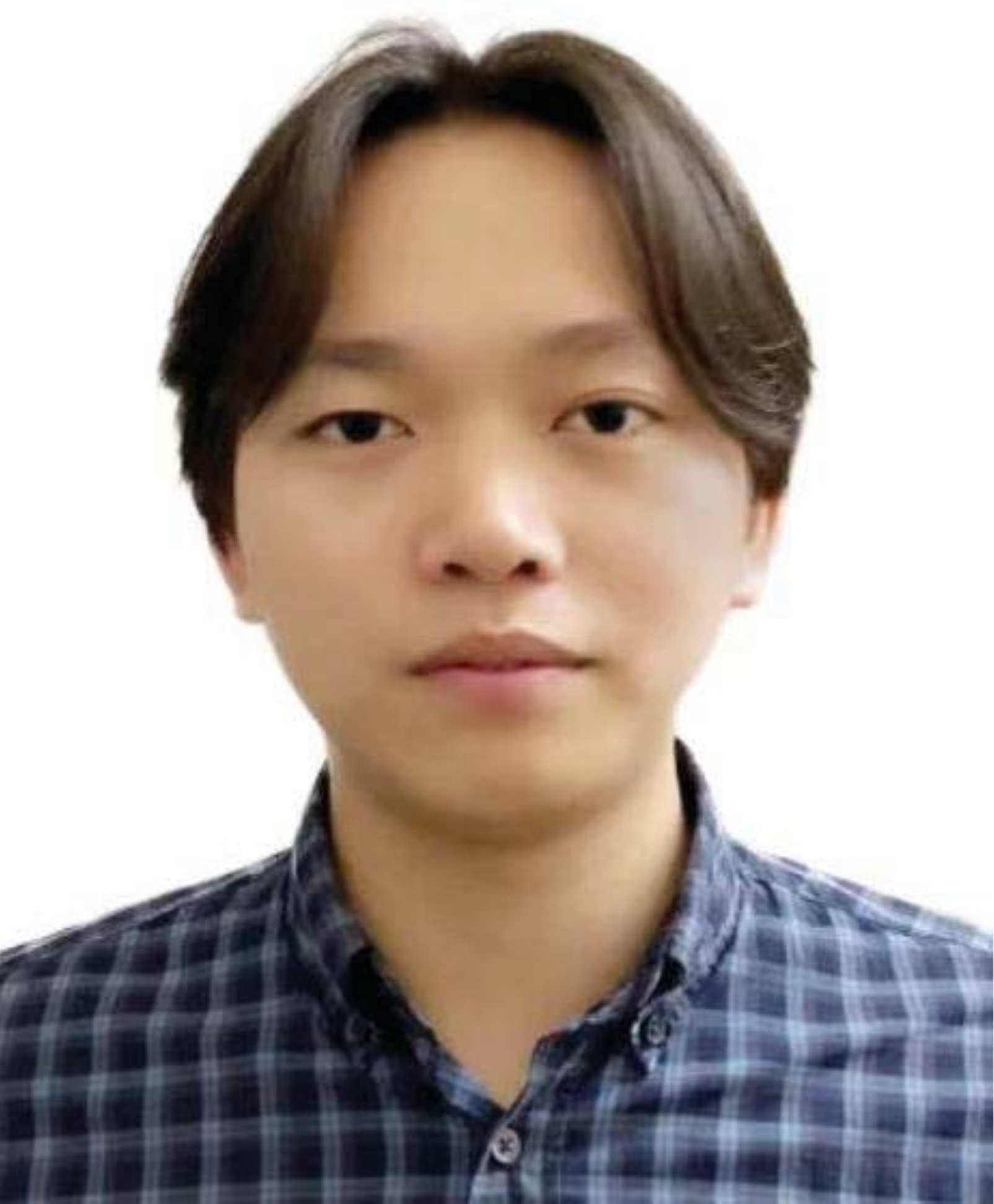}}]{Shihua Huang}
received the B.Eng. degree from Northeastern University, Shenyang, China, in 2018. 
He is currently a Research Assistant with the Department of Computer Science and Engineering, Southern University of Science and Technology, Shenzhen, China. His current research interests include representation learning, multiobjective optimization, and their applications.
\end{IEEEbiography}
\begin{IEEEbiography}[{\includegraphics[width=1in,height=1.2in,clip,keepaspectratio]{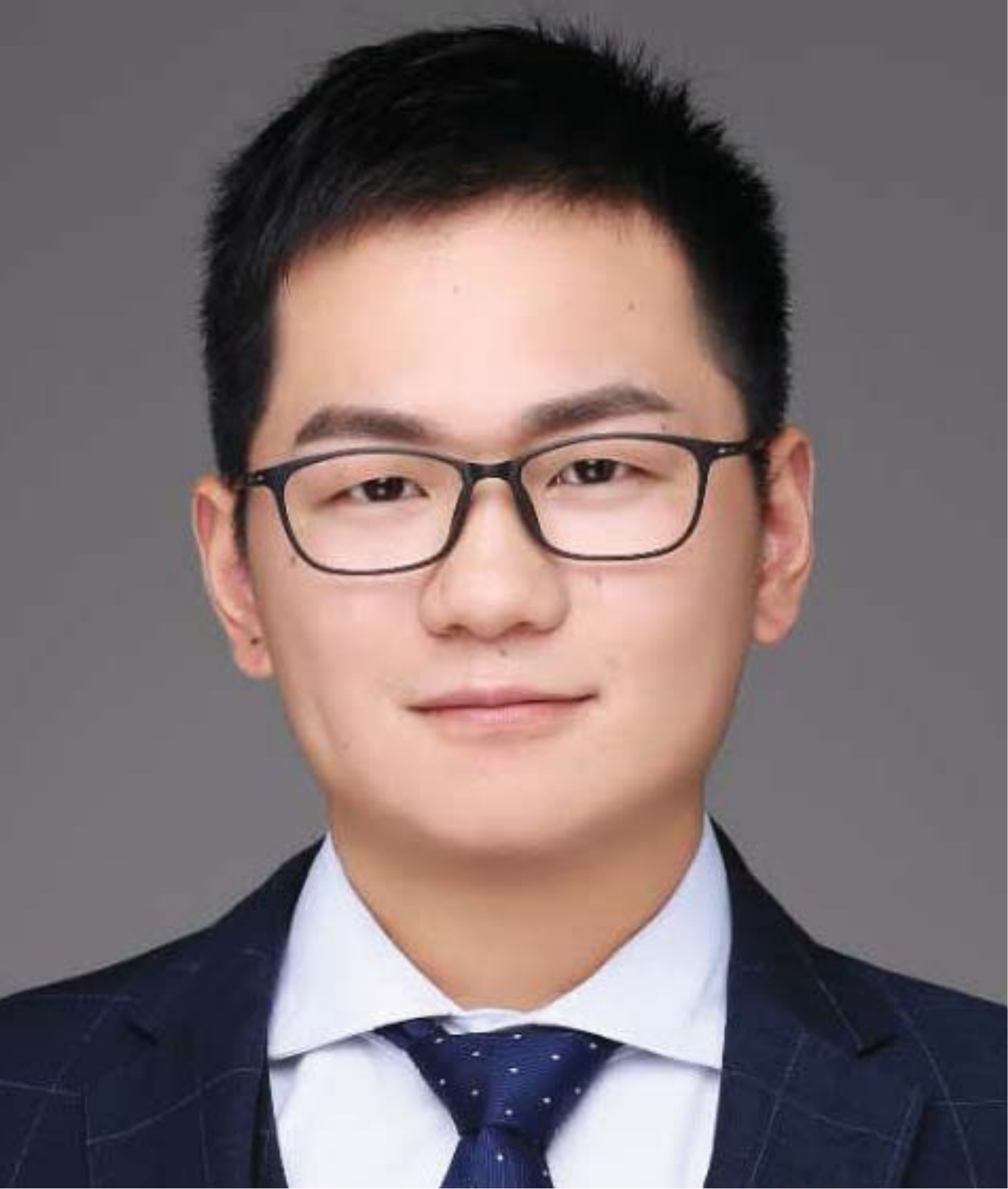}}]{Cheng He} (Member, IEEE)
received the B.Eng. degree from the Wuhan University of Science and Technology, Wuhan, China, in 2012, and the Ph.D. degree from the Huazhong University of Science and Technology, Wuhan, China, in 2018.
He is currently a Research Assistant Professor with the Department of Computer Science and Engineering, Southern University of Science and Technology, Shenzhen, China. His current research interests include model-based evolutionary algorithms, multiobjective optimization, large-scale optimization, deep learning, and their applications.
He is a recipient of the SUSTech Presidential Outstanding Postdoctoral Award from Southern University of Science and Technology, and the leading guest editor for ``SPECIAL ISSUE: Emerging Topics in Evolutionary Multiobjective Optimization'' of the Complex \& Intelligent Systems.
\end{IEEEbiography}
\begin{IEEEbiography}[{\includegraphics[width=1in,height=1.2in,clip,keepaspectratio]{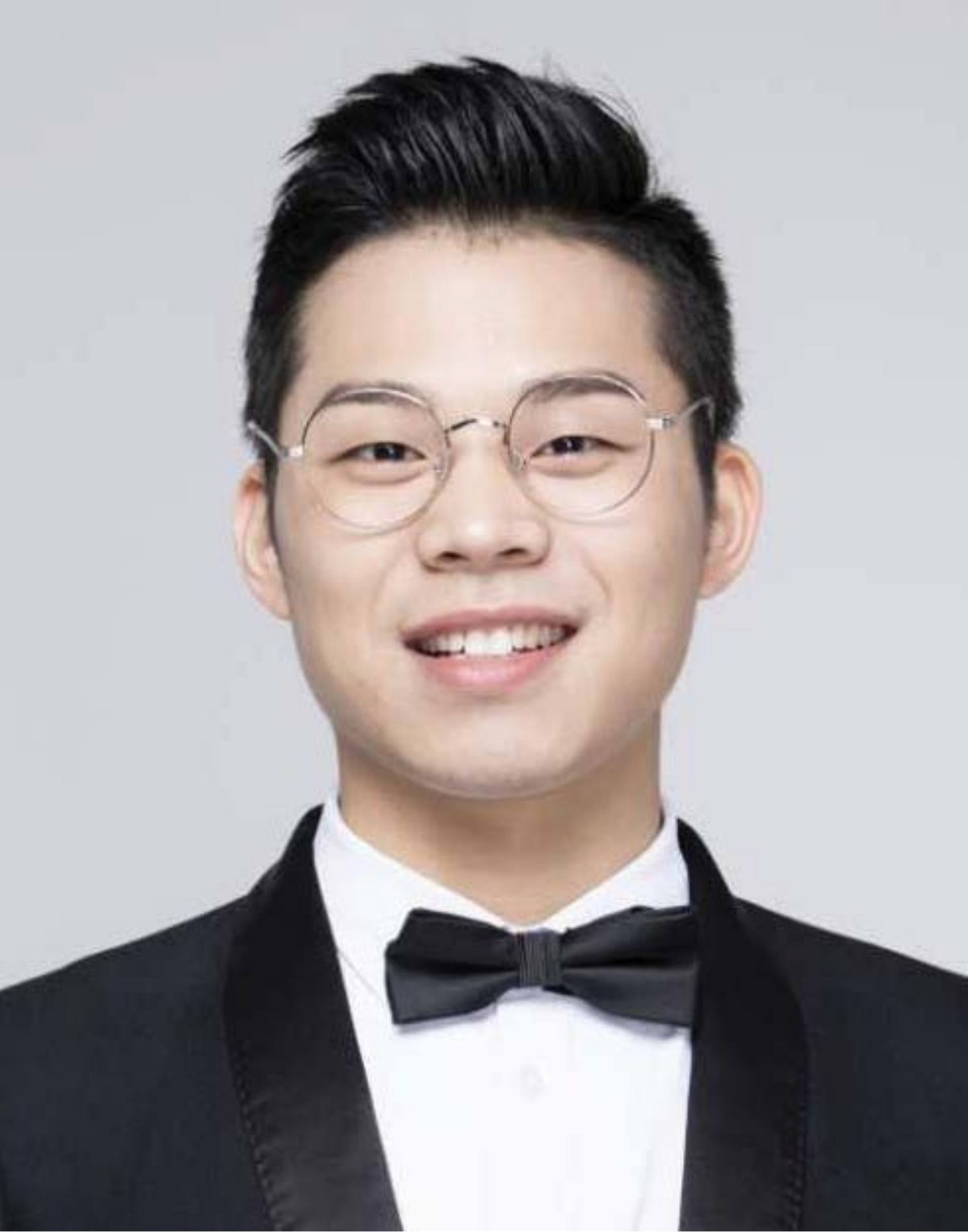}}]{Changxiao Qiu}
received the B.S. degree in 2012, and the M.S. degree in 2015 from the University of Electronic Science and Technology of China(UESTC), Chengdu, China.
He is currently a Senior Engineer with the Hisilicon Research Department, Huawei Technologies Co., Ltd., Shenzhen, China. His research interests include deep learning and computer vision.
\end{IEEEbiography}
\begin{IEEEbiography}[{\includegraphics[width=1in,height=1.2in,clip,keepaspectratio]{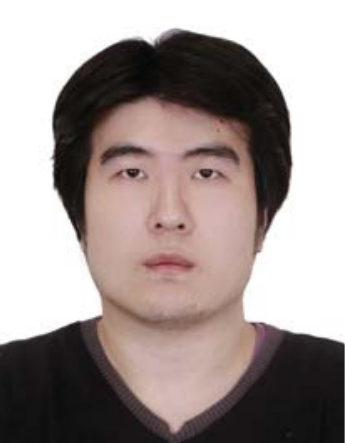}}]{Fan Yang}
received the electronic bachelor and master degree from the Paris-Sud University (University of Paris XI), 91400 Orsay, France, and the Ph.D. degree in informatics from the Paris-Saclay University, 91400 Orsay, France, in 2015.
He is currently a Principal Engineer and Project Manager with the Hisilicon Research Department, Huawei Technologies Co., Ltd., Shenzhen, China. His current research interests include neural network compression and acceleration.
\end{IEEEbiography}
\begin{IEEEbiography}[{\includegraphics[width=1in,height=1.2in,clip,keepaspectratio]{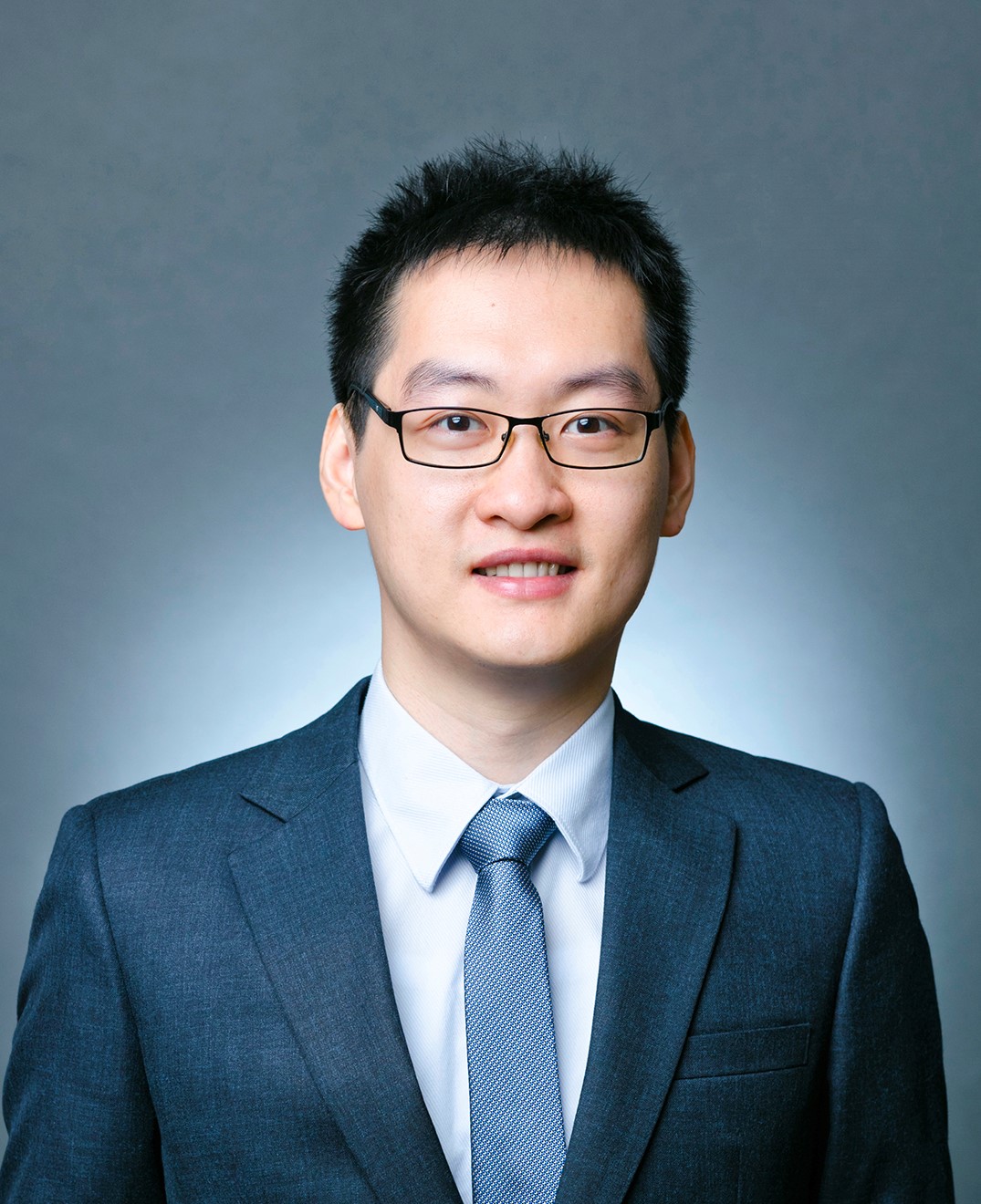}}]{Ping Luo}
is an Assistant Professor in the department of computer science, The University of Hong Kong (HKU). 
He received his PhD degree in 2014 from Information Engineering, the Chinese University of Hong Kong (CUHK), supervised by Prof. Xiaoou Tang and Prof. Xiaogang Wang. He was a Postdoctoral Fellow in CUHK from 2014 to 2016. 
He joined SenseTime Research as a Principal Research Scientist from 2017 to 2018. 
His research interests are machine learning and computer vision. He has published 100+ peer-reviewed articles in top-tier conferences and journals such as TPAMI, IJCV, ICML, ICLR, CVPR, and NIPS. His work has high impact with 13000 citations according to Google Scholar. He has won a number of competitions and awards such as the first runner up in 2014 ImageNet ILSVRC Challenge, the first place in 2017 DAVIS Challenge on Video Object Segmentation, Gold medal in 2017 Youtube 8M Video Classification Challenge, the first place in 2018 Drivable Area Segmentation Challenge for Autonomous Driving, 2011 HK PhD Fellow Award, and 2013 Microsoft Research Fellow Award (ten PhDs in Asia).

\end{IEEEbiography}

\end{document}